\documentclass{article}

    \PassOptionsToPackage{numbers, compress}{natbib}

\usepackage{enumitem} 
\usepackage[normalem]{ulem}
\newcommand{\batalg}{ALBAT}
\usepackage{amsmath}

\DeclareMathOperator*{\argmin}{arg\,min}

\usepackage{amssymb}
\usepackage{wrapfig}

\newcommand{\E}{\mathbb{E}}
\newcommand{\R}{\mathbb{R}}
\newcommand{\vtheta}{\mathbf{\theta}}

\newcommand{\vp}{\mathbf{p}}

\newcommand{\vu}{\mathbf{u}}

\newcommand{\vx}{\mathbf{x}}
\newcommand{\vy}{\mathbf{y}}

\newcommand{\mA}{\mathbf{A}}
\newcommand{\mB}{\mathbf{B}}

\newcommand{\mG}{\mathbf{G}}
\newcommand{\mH}{\mathbf{H}}
\newcommand{\mI}{\mathbf{I}}

\newcommand{\mQ}{\mathbf{Q}}

\newcommand{\mS}{\mathbf{S}}

\newcommand{\mW}{\mathbf{W}}
\newcommand{\mX}{\mathbf{X}}
\newcommand{\mY}{\mathbf{Y}}

\newtheorem{theorem}{Theorem}[section]
\newtheorem{proposition}[theorem]{Proposition}

\newtheorem{definition}[theorem]{Definition}

\usepackage{algorithm}
\usepackage{algpseudocode}

\newcommand{\ie}{\emph{i.e.}\@ifnextchar.{\!\@gobble}{}}
\newcommand{\eg}{\emph{e.g.}\@ifnextchar.{\!\@gobble}{}}

\usepackage{graphicx}
\usepackage{multirow}

\usepackage[preprint]{neurips_2025}

\usepackage[utf8]{inputenc} 
\usepackage[T1]{fontenc}    
\usepackage{hyperref}       
\usepackage{url}            
\usepackage{booktabs}       
\usepackage{amsfonts}       
\usepackage{nicefrac}       
\usepackage{microtype}      
\usepackage{mathtools}
\usepackage[table,xcdraw]{xcolor}

\title{Backbone Augmented Training for Adaptations}

\author{%
  Jae Wan Park, Junhyeok Kim, Youngjun Jun, \textbf{Hyunah Ko, Seong Jae Hwang}\\
  Department of Artificial Intelligence\\
  Yonsei University\\
  Seoul, Korea \\
  \texttt{\{paneah, timespt, youngjun, kha9867, seongjae\}@yonsei.ac.kr} \\
}

\begin{document}

\maketitle

\begin{abstract}

Adaptations facilitate efficient training of large backbone models, including diffusion models for image generation and transformer-based language models. While various adaptation techniques enhance performance with minimal computational resources, limited adaptation data often lead to challenges in training.
To address this, we focus on the enormous amount of backbone data used to pre-train the backbone models. We propose \textbf{B}ackbone \textbf{A}ugmented \textbf{T}raining (BAT), a method that leverages backbone data to augment the adaptation dataset.
First, we formulate and prove two mathematical key propositions: one establishes the validity of BAT, while the other identifies a condition under which BAT benefits adaptation.
Furthermore, we introduce an advanced data selection scheme that satisfies these propositions and present \textbf{AL}gorithm for \textbf{BAT} (\batalg{}) to implement this approach. \batalg{} efficiently enhances adaptation training in both personalization and language generation tasks with scarce data.

\end{abstract}
\section{Introduction}
\label{sec:intro}

Recently, large foundation models have demonstrated exceptional performance across various tasks \cite{gpt3, ldm, llama3, dit, sd3}. To adapt these models for specific downstream tasks, a variety of adaptation techniques have been introduced. These approaches typically involve updating only a small portion of parameters from  the large model, often called backbone models.
Some have leveraged rank decomposition \citep{lora, qlora, dora} of the backbone weights, while others employing fixed text embeddings \citep{dreambooth, text_inversion} to maintain identity consistency in image generation.

Adaptation methods have enabled any user, equipped with an appropriate adaptation dataset, to fine-tune large models for tasks defined by themselves.
However, without sufficient data, the adaptation is likely to suffer from various complications.
Primarily, underfitting may occur, and at the same time, efforts to alleviate underfitting by increasing the number of training steps can result in overfitting \citep{overfitting1, overfitting2, overfitting3}.
Thus, obtaining sufficient amount of \textit{adaptation data} is crucial, but imposes a significant burden on users \citep{data1, data2, data3}.

Meanwhile, recent studies have highlighted the importance of the distribution of backbone  \citep{aligning} in dealing with data scarcity. There has been a work that utilized pre-train dataset in language model~\cite{prior2}, then in vision tasks~\cite{prior1}. In adaptations, DreamBooth \citep{dreambooth} utilizes synthetic data generated by the backbone model for regularization, while DoRA \citep{dora} proposes an adaptation method that aligns training behavior with that of the backbone. Inspired by these approaches, we propose a method called Backbone Augmented Training (BAT) to mitigate data scarcity. BAT augments adaptation data by leveraging the \textit{backbone data} used during the pre-training of the backbone model.



Although approaches similar to BAT have recently been proposed within the community \cite{civitai}, they perform augmentation using backbone data without any theoretical validation. For instance, \cite{civitai} arbitrarily selects backbone data that appear likely to enhance adaptation performance. Consequently, while these methods occasionally yield favorable results, they often perform worse than cases without augmentation. To overcome these fundamental flaws, we establish a mathematical foundation of BAT for the first time, demonstrating its theoretical validity. 

\begin{wrapfigure}{r}{0.6\textwidth}
    \centering
    \vspace{-15pt}
    \includegraphics[width=\linewidth]{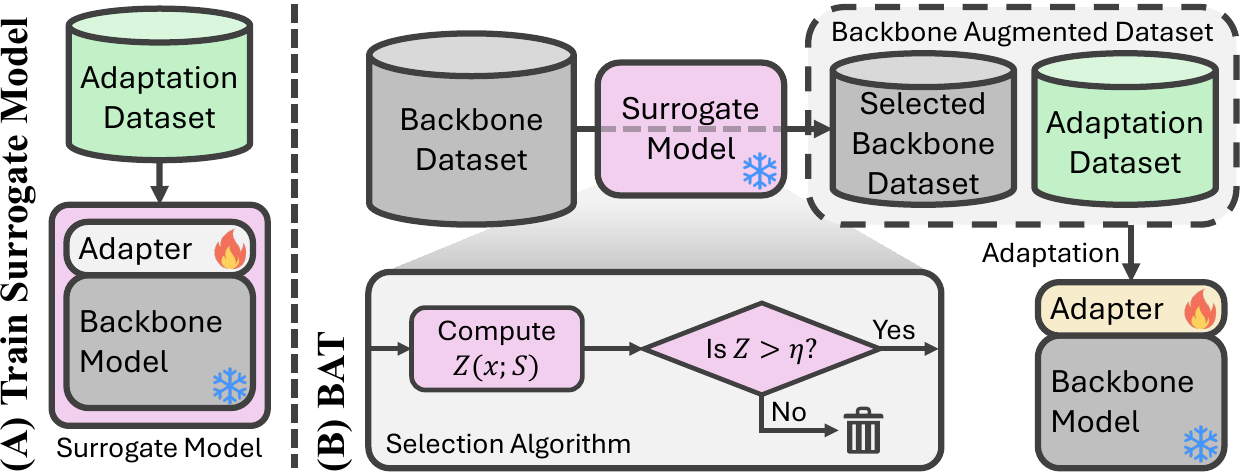}
    \caption{\textbf{Framework of our Backbone Augmented Training.} (A) The surrogate model is first trained using the adaptation dataset, while the backbone model is pre-trained separately. (B) With the help of the surrogate model, backbone data are selected using a selection scheme grounded in our solid mathematical foundation. This approach differentiates itself from traditional adaptation methods that do not use backbone data and existing methods \cite{civitai} that arbitrarily select backbone data.}
    \label{fig:pipeline-fig}
\end{wrapfigure}

To establish a solid mathematical foundation for BAT, we begin by adopting reasonable assumptions proposed in \cite{textbook}. 
Building on these assumptions, we derive two key propositions to verify the rationale and efficiency of BAT.
The first proposition demonstrates that BAT allows the adaptation parameters to converge to the optimal parameters, justifying the use of backbone data in adaptations.
The second proposition offers a fundamental condition that helps to monitor the convergence rate of BAT.
According to the second proposition, BAT outperforms traditional adaptation methods \citep{lora,dreambooth,dora} when the selected backbone data satisfy specific conditions.

An effective strategy for selecting backbone data that meets the condition of our second proposition necessarily facilitates adaptation training with BAT. Drawing inspiration from data selection studies under weak supervision \citep{keypaper}, we propose a biased selection scheme that thoroughly selects a subset of the backbone data to augment the adaptation dataset. 
While the biased scheme has been shown to be more effective than random selection or an unbiased scheme \citep{keypaper}, it faces the challenge of high impractical computational costs due to Hessian derivatives.
We address this limitation by adopting mathematical methodologies inspired by \citep{datainf}, reducing the computational cost from $O(D^3)$ to $O(D)$, for $D$ adaptation parameters.
Finally, we devise a complete pipeline, referred to as \batalg{} (Fig.~\ref{fig:pipeline-fig}), which implements this selection scheme.

\begin{figure*}[ht!]
    \begin{center}
    \centerline{\includegraphics[width=\textwidth]{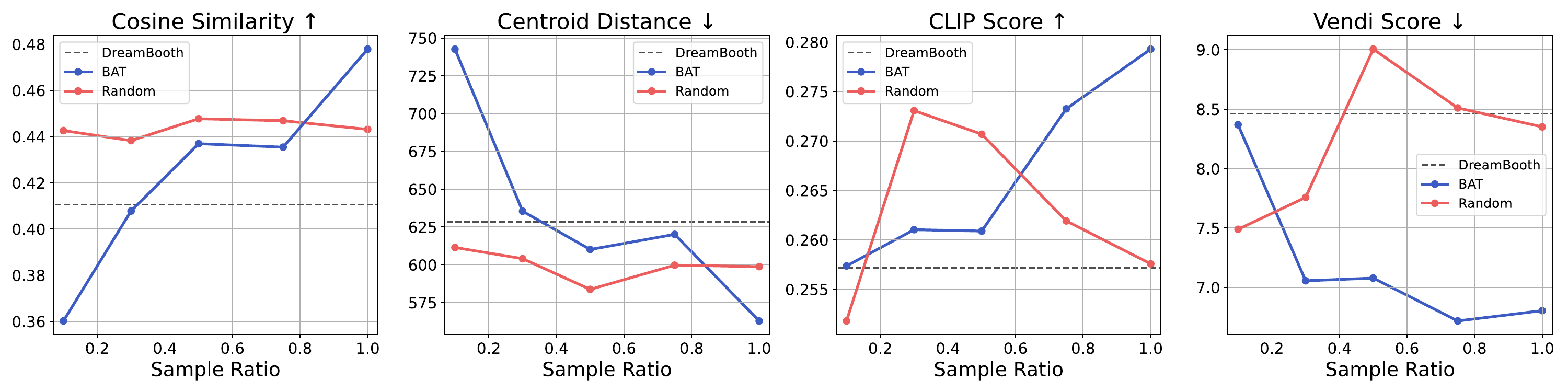}}
    \caption{
    \textbf{Application Results of BAT in Personalized Image Generation Task.} The comparison is conducted between two approaches: augmenting a text-to-image model personalized via DreamBooth \cite{dreambooth} with arbitrary selected backbone data (Random) and utilizing our BAT data selection scheme (BAT). The sample ratio represents the proportion of backbone dataset used for data selection while equal amount of backbone data is selected for each sample ratio. Each metric evaluates the performance of personalized image generation. While the random approach shows better performance only for few metrics and no correlation with the sample ratio, BAT exhibits consistent performance improvement as the sample ratio increases, achieving results that surpass others and coinciding with data selection theory \citep{keypaper}. For more details, please refer to Sec.~\ref{ssec:validation}.
    }
    \label{fig:result1}
    \end{center}
    \vskip -0.2in
\end{figure*}

To sum up, our main contributions are as follows:
\begin{itemize}[left=0pt]
    \item As a method to augment the adaptation dataset, we propose Backbone Augmented Training (BAT). Through Propositions~\ref{prop:p1} and \ref{prop:p2}, we establish the theoretical foundation of BAT and define a condition under which it is effective. 

    \item Building on our mathematical foundation, we propose \batalg{}, an algorithm for selecting useful backbone data. 
    To this end, we make data influence computation feasible by integrating multiple mathematical techniques.

    \item Beyond theoretical validation, we conduct experiments using \batalg{} on personalized image generation (Fig.~\ref{fig:result1}) and language generation tasks (Table~\ref{tab:tab1}) in data-scarce scenarios, demonstrating that applying BAT improves adaptation performance.
\end{itemize}
\section{Related Works}
\label{sec:related}

\textbf{Adaptations.} Fine-tuning a large-scale model to solve a downstream task is extremely expensive. To mitigate this challenge, many have come up with methods that train a small portion of parameters, also known as adaptations. Adaptation methods are widely distinguished as additive fine-tuning \citep{serialadapter, prefixtuning}, selective fine-tuning 
\citep{bitfit, diffpruning}, reparameterized fine-tuning \citep{intrinsic, compacter}. In the following part, we introduce eminent types of adaptations.

\textbf{LoRA.} Low-Rank Adaptation~\citep{lora} has gained significant attention among early adaptations for its ability to efficiently train a small portion of parameters through weight decomposition, without any additional inference burden. Despite utilizing only a small set of parameters, LoRA achieves performance comparable to full fine-tuning, and in certain benchmarks, even surpasses it. Based on the strong performance of LoRA, several variants including DoRA~\citep{dora, qlora} emerged for language models. Others applied this decomposition method in generative models such as diffusion models~\citep{ldm, ddim, ddpm} like LoCon, LoHA and LoKr~\citep{lycoris}. While LoRA-based approaches have demonstrated significant improvements in training efficiency, the lack of adaptation data remains as a challenge and often leads to overfitting.

\textbf{DreamBooth.} DreamBooth \citep{dreambooth} aims to generate images of a specific subject only with the reference images of that subject, using text-to-image (T2I) diffusion models. 
Concretely, given images of a subject, DreamBooth learns to bind that specific subject with a unique token identifier. Not only to generate the target subject effectively but also to preserve the knowledge of the pre-trained model, DreamBooth utilizes synthesized images as a regularization dataset.
However, for specialized tasks, gathering the adaptation dataset required to train Dreambooth is extremely challenging, placing a significant burden on users.

\textbf{Data Selection.}
Data selection remains a vital and actively researched topic in model training \citep{data_selection, data_selection2, data_selection3}. Often using empirical risk minimization setting, the risk is defined as $R(\theta) = \sum^{n}{S_i\mathcal{L}_i}$, where $\mathcal{L}_i$ represents the loss for each data point. This formulation allows it to control the influence of individual data points through $S_i$. Several studies \citep{influence, influnce2, leverage} define $S$ as an unbiased scheme when $\E[S(x_i)|x_i] = 1$, and as a biased scheme otherwise. Notably, \cite{keypaper} has shown that the biased schemes surpass random or unbiased schemes in lowering the empirical risk. However, biased schemes require immense amount of calculation particularly for deep learning, resulting in no practical application to modern models.
To address this, we leverage various mathematical techniques from prior studies \citep{damping, grosse, datainf} to reduce the computational cost to a feasible level for current deep learning applications.


\section{Definitions}
\label{sec:back}

To formulate Backbone Augmented Training mathematically, we introduce the notations and definitions consistently used throughout the subsequent sections. For conciseness, we review notations commonly employed in probabilistic asymptotics (Supp.Sec.~\ref{ssec:notation}). Then, in the following, we define the mathematical definition of adaptation (Sec.~\ref{ssec:mathematics}) and provide the definitions essential for discussing the backbone augmentation scheme (Sec.~\ref{ssec:definitions}).


\subsection{Mathematics on Adaptations}
\label{ssec:mathematics}

Every adaptation method begins by initializing parameters from its backbone model.
Let $\vtheta^\text{B} \in \Theta^\text{B}$ represent the backbone model parameters, and $\vtheta^\text{A} \in \Theta^\text{A}$ encompass all parameters from both the backbone model and adapter.
Then, loading an initialized adapter over the backbone model can be expressed using a continuous function $g$, such that $\vtheta^\text{A} := g(\vtheta^\text{B}) \in \Theta^\text{A}$. That is, denoting $\vtheta^\text{A} \backslash \vtheta^\text{B}$ as the parameters exclusive to the adaptation, note that $0 < \text{dim}(\vtheta^\text{A} \backslash \vtheta^\text{B}) < \text{dim}(\vtheta^\text{B})$ holds.
Typically, the backbone model itself is frozen, allowing only a small subset of parameters to be updated.
Thus, as the training step $n$ progresses and the $\hat\vtheta_{n}^\text{A}$ are updated to reach their optimal values $\vtheta^{\text{A}^{*}}$, the parameter training is described as $\textstyle {(\hat\vtheta^\text{A}\backslash \vtheta^\text{B})}_{n+1} = {(\hat\vtheta^\text{A}\backslash \vtheta^\text{B})}_{n} + \Delta(\vtheta^\text{A} \backslash \vtheta^\text{B})_{n+1}$.

We now define risk functions. To begin, let the backbone model be pre-trained with the dataset $\displaystyle \mathcal{D}^\text{B}$ via empirical risk minimization. Suppose dataset $\displaystyle \mathcal{D}^\text{A}$ is a training set for adaptation, usually constructed by the user. The size of each dataset is noted as $\displaystyle N^B:= |\mathcal{D}^\text{B}|$ and $\displaystyle N^A:=|\mathcal{D}^\text{A}|$, respectively, where $\displaystyle n \ll N$ is common due to the nature of adaptation.
Recall that backbones and adaptations commonly share the loss function.
We set the backbone risk $R^\text{B}_N$ as
\begin{equation} 
\begin{aligned}
     R^\text{B}_{N^B} &:= \frac{1}{N^B}\sum_{\vx, \vy \in \mathcal{D}^\text{B}}\mathcal{L}(\vy, f^\text{B}(\vx;\vtheta^\text{B})) + \lambda\Omega(\vtheta^\text{B}), \\
     \displaystyle \vtheta^{\text{B}^{*}} &:= \underset{\Theta^\text{B}}{\argmin}~R^\text{B}_{N^B},
\end{aligned}
\end{equation}
with the regularizer function $\displaystyle \Omega: \Theta \rightarrow \R$ and constant $\displaystyle \lambda$ to balance the training. On the other hand, adaptation risk $R^\text{A}_n$ is defined as
\begin{equation} 
\begin{aligned}
    R^\text{A}_{N^{A}} &:= \frac{1}{{N^{A}}}\sum_{\vx, \vy \in \mathcal{D}^\text{A}}\mathcal{L}(\vy, f^\text{A}(\vx;\vtheta^\text{A})) + \lambda\Omega(\vtheta^\text{A}), \\
    \displaystyle \vtheta^{\text{A}^{*}} &:= \underset{\Theta^\text{A}}{\argmin}~R^\text{A}_{N^{A}}.
\end{aligned}
\end{equation}
Since $\displaystyle \mathcal{D}^\text{B} \cap \mathcal{D}^\text{A} = \varnothing$, the adaptation risk possesses independent characteristics from the backbone risk. Thus, combining both risks does not necessarily ensure convergence.

\subsection{Definitions of Backbone Augmented Training}
\label{ssec:definitions}
Now, we construct the essential definitions to mathematically formalize Backbone Augmented Training.
\begin{definition}
\textit{Backbone augmented training risk on an adaptation is defined as}
\label{def:composite_risk}
\begin{equation} 
    R^{\text{bat}|\text{A}}_k := \frac{1}{k}\sum_{\vx, \vy \in \mathcal{D}^{\text{bat}|\text{A}}}\mathcal{L}(\vy, f^\text{A}(\vx;\vtheta^{\text{bat}|\text{A}})) + \lambda\Omega(\vtheta^{\text{bat}|\text{A}}),
    \nonumber
\end{equation}
\textit{for some} $\displaystyle \mathcal{D}^{\text{bat}|\text{A}} = \mathcal{D}^{\text{B}^\prime} \cup \mathcal{D}^\text{A}$ \textit{where} $\varnothing \neq \mathcal{D}^{\text{B}^\prime} \subset \mathcal{D}^\text{B}.$ \textit{Here,} $k = |\mathcal{D}^{\text{bat}|\text{A}}|$ \textit{denotes the cardinality of} $\mathcal{D}^{\text{bat}|\text{A}},$ \textit{and} $\hat\vtheta_1^{\text{bat}|\text{A}} = \hat\vtheta_1^\text{A}$ \textit{is assumed for the first training step.}
\end{definition}

The notation $\text{bat}|\text{A}$ in Def.~\ref{def:composite_risk} stands for the application of BAT to the adaptation $\text{A}$. Specifically, the adaptation's training set $\mathcal{D}^\text{bat}$ incorporates not only the original adaptation dataset $\mathcal{D}^\text{A}$ but also a portion of backbone dataset $\mathcal{D}^\text{B}$ (i.e., $\mathcal{D}^{\text{B}^\prime}$).
Meanwhile, $\displaystyle \hat\vtheta_1^{\text{bat}|\text{A}} = \hat\vtheta_1^\text{A}$ indicates that when BAT training for an adaptation begins, the adaptation with BAT is initialized with the same weights as the standard adaptation.
We demonstrate in the following section that this BAT risk always increases the performance of adaptations when specific condition (Proposition~\ref{prop:p2}) is satisfied, unlike other random mixture of backbone and adaptation risks.

\begin{definition}
\label{def:def2}
\textit{Backbone augmentation ratio is denoted as} \( \displaystyle n/k \rightarrow \gamma \in (0, 1). \)
\end{definition}

This ratio essentially shows the proportion of adaptation data and backbone data used in our method. In this definition, we use convergence to derive the ratio and adopt it in our proposition based on asymptotic.

Lastly, following the format of prior studies on estimators \citep{textbook, keypaper}, we extend our discussion by introducing asymptotic error coefficients.
We begin by defining the weighted quadratic error as $\displaystyle ||\hat\vtheta - \vtheta^*||^2_\mS := \langle \hat\vtheta - \vtheta^*, \mS(\hat\vtheta - \vtheta^*) \rangle$, for some $\displaystyle \mS \in \R^{\text{dim}(\Theta) \times \text{dim}(\Theta)}$. Notably, setting $\displaystyle S=\mI$ gives a simple Euclidean inner product  when $\displaystyle R_N$ is twice differentiable.
Additionally, choosing $\displaystyle \mS = \nabla^2_{\vtheta}R_N$ results in the total risk achieved from the iteration through entire epoch of $\displaystyle \mathcal{D}^\text{B}$ (please refer to~\cite{keypaper} for more detailed structure).
Finally, we define the asymptotic error coefficient as $\displaystyle \rho_{_\text{B}}(\mS) := {p\!\!-\!\!\lim}_{{N^{B}} \rightarrow \infty}N||\hat\vtheta^\text{B} - \vtheta^{\text{B}^*}||^2_\mS$. 
\begin{definition}
\label{def:bac}
The \textit{backbone augmentation coefficient on an adaptation is defined as}
\begin{equation} 
    \rho_{_{\text{bat}|\text{A}}}(\mS) := {p\!\!-\!\!\lim}_{k \rightarrow \infty}k||\hat\vtheta^{\text{bat}|\text{A}} - \vtheta^{\text{A}^*}||^2_\mS.
    \nonumber
\end{equation}
\end{definition}
This coefficient represents the convergence rate toward the optimal adaptation.
Its convergence depends on the limit of the estimator. If the coefficient remains a real value, the estimator is guaranteed to converge to the optimal parameters.

We further define the conditional Hessian matrix for parameters of the backbone risk as $\displaystyle \mH^\text{B}(\vx) \coloneqq \E[\nabla^2_{\vtheta}\mathcal{L}^{\vtheta^{\text{B}^*}}|\vx]$. This notation is useful in representing the parameter update in optimization with respect to related variables. If the notation $\displaystyle \text{B}$ is replaced, then the matrix is associated with another model and its empirical risk.
\section{Mathematical Foundations for Backbone Augmented Training}
\label{sec:proposition}

Using the notations established in Sec.~\ref{sec:back}, we formally validate the \textbf{B}ackbone \textbf{A}ugmented \textbf{T}raining (BAT) (Proposition~\ref{prop:p1}), and identify the conditions under which BAT demonstrates its effectiveness (Proposition~\ref{prop:p2}).


Proposition \ref{prop:p1} is mainly about the convergence of the backbone augmentation training. Since the coefficient $\displaystyle \rho_{_{\text{bat}|\text{A}}}$ represents the rate of convergence, the existence of $\displaystyle \rho_{_{\text{bat}|\text{A}}}$ implies that BAT adaptation will eventually converge to its optimal parameters.
Utilizing this proposition, we justify BAT specifically in DreamBooth \citep{dreambooth} and LoRA \citep{lora} in Supp.~Sec.~\ref{supp-ssec:proof_adapters}. 

Proposition~\ref{prop:p2} shows the basic condition for backbone data to surpass the regular adaptation training.
If we select data in $\mathcal{D}^{\text{bat}|\text{A}}$ that satisfy this inequality, BAT will surpass the regular adaptation training.
This also explains why some arbitrary backbone augmentations have occasionally yielded favorable results---because the selected backbone data largely adhered to Proposition~\ref{prop:p2}.

\subsection{Assumptions}
\label{ssec:assumptions}

For conciseness, we revisit four assumptions (\textit{A1}, \textit{A2}, \textit{A3}, \textit{A4}) that are basic in asymptotic estimation theories \citep{keypaper} in the scope of the nature of the backbone and adaptation risks in Supp.Sec.~\ref{assumptions}. Then, we propose another novel assumption (\textit{A5}) to solidify the property of our method's risk, as the risk is originally proposed in this work.

\begin{enumerate}[itemsep=0pt]
    \item[\textit{A5.}] \textit{For any neighborhood} $\displaystyle U^n$ \textit{of} $\displaystyle \vtheta^{\text{A}^*}$ \textit{where} $\displaystyle \hat\vtheta_n^{\text{bat}|\text{A}} \in U^n$, \textit{it follows that:}
    \begin{equation} 
        R^\text{A}(\vtheta) - R^{\text{bat}|\text{A}}(\vtheta) \neq R^\text{A}(\vtheta^{\text{A}^*}) - R^{\text{bat}|\text{A}}(\vtheta^{\text{A}^*}; \mathcal{D}^{\text{bat}|\text{A}}),
        \nonumber
    \end{equation}
    \textit{for any} $\displaystyle \vtheta\in \Theta^\text{A}$ \textit{except} $\displaystyle \vtheta = \vtheta^{\text{A}^*}.$
\end{enumerate}
\textit{A1} asserts that the risks have unique minimum values, which is a standard assumption in theoretical proofs \citep{keypaper, formestudy1}. \textit{A2} ensures that the risks are continuous and finite. \textit{A3} and \textit{A4} assume that both the backbone and adaptation risks are differentiable and convex.
Finally, \textit{A5} presumes that the composite risk defined in our method is a smooth function in a neighborhood containing the adaptation's optimal parameter.

\subsection{Main Propositions}
\label{sec:propositions}

Upon the assumptions in Sec.~\ref{ssec:assumptions}, we present two propositions regarding our method's risk. Due to the limited space, we leave the proofs in Supp.~Sec.~\ref{supp-ssec:p1_proof} and Supp.~Sec.~\ref{supp-ssec:p2_proof}.

\begin{proposition}
    \label{prop:p1}
     \textbf{(Validity of BAT)} \\ \textit{Suppose the assumptions in Sec.~\ref{ssec:assumptions} hold. Then, for any} $\displaystyle \mS \in \R^{dim(\Theta^\text{A}) \times dim(\Theta^\text{A})}$ \textit{that is symmetric,}
$\displaystyle \rho_{_{\text{bat}|\text{A}}}(\mS)$ \textit{exists.} 
\end{proposition}

\begin{proposition}
    \label{prop:p2}
    \textbf{(Condition for BAT)} 
    \vspace{3pt}
    \\
    \textit{Let} $\mathcal{D}^{\text{bat}|\text{A}} \cap \mathcal{D}^\text{B}  = \mathcal{D}^{\text{B}^\prime}$\textit{, and} $\mH^\text{bat} = \E[\nabla^{2}_{\vtheta} \mathcal{L}^{\text{bat}|\text{A}}|\vx]$ $\displaystyle \iff (\vx, \vy) \in \mathcal{D}^{\text{B}^\prime}$. \textit{If the following condition,}
    \begin{equation} 
    \begin{aligned}
        \gamma &||(\mH^{\text{bat}|\text{A}})^{-1}\sum_{\mathcal{D}^\text{bat}}\nabla_\vtheta\mathcal{L}^{\text{bat}|\text{A}}|| \le
        ||(\mH^{\text{bat}|\text{A}} - \mH^\text{bat})^{-1}\sum_{\mathcal{D}^\text{A}}\nabla_\vtheta\mathcal{L}^{\text{bat}|\text{A}}|| + o_P(1),
    \end{aligned}
    \nonumber
    \end{equation}
    \textit{holds for any} $\scriptstyle \vtheta \in (\vtheta^\text{A} \cap \vtheta^\text{B})$, \textit{then} $\displaystyle \rho_{_{\text{bat}|\text{A}}} \le \rho_{_\text{A}}$\textit{holds under the assumptions of Proposition~\ref{prop:p1}. Moreover, unless} $\displaystyle \gamma \rightarrow 1,$ \textit{the inequality is strict.}
\end{proposition}


Note that throughout the proof of Proposition~\ref{prop:p2}, $\displaystyle \mH^\text{A}$ vanishes, indicating that the Hessian calculation for the original adaptation is no longer necessary. Therefore, tracking only $\displaystyle \mH^{\text{bat}|\text{A}}$ is sufficient, reducing memory complexity. However, since all Hessian calculations are inherently computationally expensive, the criterion of Proposition~\ref{prop:p2} still remains impractical for applications on deep learning models.
For DreamBooth, it requires 280.24 GB of VRAM and LoRA with LLaMA 3-B needs 384.16GB of VRAM to calculate this equation. Since common GPUs have VRAM of 24 to 40GB, this is surely a computational burden.
To address these challenges, in Sec.~\ref{sec:scheme}, we introduce a data selection scheme that not only satisfies Proposition~\ref{prop:p1} but also reduces time and memory complexity drastically: 16.74GB for DreamBooth and 19.65GB for LoRA.




\section{Algorithm for Backbone Augmented Training}
\label{sec:scheme}
In this section, we solidify a data selection scheme upon our theoretical foundation (Sec.~\ref{sec:selection_scheme}), and develop an algorithm that effectively selects backbone data (Sec.~\ref{sec:selection_algorithm}).
To the best of our knowledge, this work is the first to present a biased backbone data selection algorithm that operates within commonly available computational resources.

\subsection{Selection Scheme}
\label{sec:selection_scheme}
We aim to propose a data selection scheme that meets our propositions while maintaining a practically feasible memory complexity. To begin, we redefine the BAT risk as:
\begin{equation} 
\begin{aligned}
    R^{\text{bat}|\text{A}}_k := \frac{1}{k}&\sum_{\vx, \vy \in \mathcal{D}^\text{B}}S(\vx)\mathcal{L}(\vy, f^\text{A}(\vx;\vtheta^{\text{bat}|\text{A}})) 
    + \sum_{\vx, \vy \in \mathcal{D}^\text{A}}\mathcal{L}(\vy, f^\text{A}(\vx;\vtheta^{\text{bat}|\text{A}})) + \lambda\Omega(\vtheta^{\text{bat}|\text{A}}),
\end{aligned}
\end{equation}
where $S(\vx) \coloneqq \mathbb{P}((\vx, \vy) \in \mathcal{D}^{\text{bat}|\text{A}})$ denotes a selection scheme.
Notably, biased selection schemes (i.e., $\E[S(x_i)|x_i] \neq 1$) have been shown to outperform both random selection and standard influence functions, which are implicitly unbiased \citep{keypaper}.
Thus, we use Portmanteau’s theorem and Radon-Nikodym derivative to derive a simplified form as our biased selection scheme, as follows\footnote{In detail, $S(x)=b(x)\in[0,1]$ if $Z(\vx;S)=\eta$. However, as this scenario does not arise in practice, we omit this zero-measure case for simplicity.}:
\begin{equation} 
\label{eq:scheme}
S(x) =
\begin{cases}
1, & \text{if } Z(\vx;S) > \eta \\
0, & \text{if } Z(\vx;S) < \eta 
\end{cases},
\end{equation}
where $\eta$ is chosen so that $\E[S(\vx)]$ = $\gamma$. The score function $Z$ is defined as follows:
\begin{equation} 
\label{eq:score_function}
\begin{aligned}
Z(\vx; S) \coloneqq
&- \operatorname{Tr}\big(\mG^{\text{bat}|\text{A}}(\vx) (\mH^{\text{bat}|\text{A}})^{-1} \mQ (\mH^{\text{bat}|\text{A}})^{-1}\big) \\
&+ 2 \operatorname{Tr}\big(\mH^{\text{bat}|\text{A}}(\vx) (\mH^{\text{bat}|\text{A}})^{-1} \mQ (\mH^{\text{bat}|\text{A}})^{-1} \mG^{\text{bat}|\text{A}} (\mH^{\text{bat}|\text{A}})^{-1}\big),
\end{aligned}
\end{equation}
where $\textstyle \mG$ is given by $\textstyle \mathbb{E}[\nabla_\theta \mathcal{L}(\theta; \vy, \vx) \nabla_\theta \mathcal{L}(\theta; \vy, \vx)^\top \vert x]$ and $\mQ$ is Hessian of the total sum of validation error. The whole derivation is in Supp. Sec. \ref{z_derivation}. According to \cite{keypaper}, this setting yields an asymptotic error coefficient $\rho$. Through our Proposition~\ref{prop:p1}, the existence of this coefficient (i.e., $\rho_{_{\text{bat}|\text{A}}}$) is guaranteed. Therefore, adopting this scheme in BAT is mathematically justified.

However, computing $Z$ under this scheme still involves calculating $\mH$, $\mG$, and $\mH^{-1}$, which requires a high time complexity of $O(nD^2L + D^3L)$, making the selection scheme impractical.
To address this, we reduce the time complexity by employing mathematical methodology inspired by DataInf~\citep{datainf}.
As a result, our selection scheme is sufficient with a time complexity of $O(nDL)$, as formalized in the following theorem.
\begin{theorem}
\label{thm:theorem}
\textbf{(Time Complexity for BAT})\\
Assume that an empirical risk is defined with a logarithm function, \( n \) data points, \( L \) model layers, and \( D \) bounding parameters for each layer. Then \( Z(\vx; S) \) is bounded by \( O(nDL) \).
\end{theorem}

\noindent
\begin{minipage}[t]{0.45\textwidth}
  \vspace{0pt} 
      \begin{algorithm}[H]
        \caption{BAT Algorithm} \label{alg:train_theta}
    \begin{small}
    \begin{algorithmic}[1] 
       \State \textbf{Input:}
       \State \hspace{1.4em} $\mathcal{D}^\text{A} \leftarrow$ Adaptation dataset
       \State \hspace{1.4em} $\mathcal{D}^\text{B} \leftarrow$ Backbone dataset
       \State \hspace{1.4em} $f(\cdot;\theta) \leftarrow$ Adapter to be trained
       \vspace{0.4em}
       \hrule
       \vspace{0.4em}
       \State Choose $\eta$ such that $\mathbb{E}[S(x)] = \gamma$
       \vspace{0.2em}
       \State $f(\cdot;\hat\vtheta^{\text{bat}|\text{A}}_\text{surrogate}) \leftarrow$ $f(\cdot;\vtheta)$ trained with $\mathcal{D}^\text{A}$
       \vspace{0.2em}
       \For{$X^\text{B}$ in $\mathcal{D}^\text{B}$}
          \State Compute $Z(X^\text{B};S)$ using $f(\cdot;\hat\vtheta^{\text{bat}|\text{A}}_\text{surrogate})$
          \If{$Z(X^\text{B};S) < \eta$}
             \State Remove $X^\text{B}$ from $\mathcal{D}^\text{B}$
          \EndIf
       \EndFor
       \vspace{0.2em}
       \State $\mathcal{D}^{\text{bat}|\text{A}} \leftarrow \mathcal{D}^\text{A} \cup \mathcal{D}^\text{B}$
       \State Train adaptation $f(\cdot;\vtheta)$ with $\mathcal{D}^{\text{bat}|\text{A}}$
    \end{algorithmic}
    \end{small}
    \vspace{-3pt}
    \end{algorithm}
\end{minipage}%
\hfill
\begin{minipage}[t]{0.5\textwidth}
  \vspace{10pt} 
    Its proof is given in Supp.~Sec.~\ref{supp-ssec:theorem_proof}. Note that Theorem~\ref{thm:theorem} can be applied to modern deep learning models such as diffusion models or transformer-based models, since they commonly adopt logarithm risk functions. According to this theorem, the time complexity has been drastically reduced.

    \subsection{Selection Algorithm}
    \label{sec:selection_algorithm}
    Building on the selection scheme in Sec.~\ref{sec:selection_scheme}, we propose \textbf{AL}gorithm for \textbf{BAT} (\batalg{}), an algorithm that selects useful backbone data using Eq.~\ref{eq:scheme}.
    The overall pipeline of \batalg{} is illustrated in Fig.~\ref{fig:pipeline-fig} and described in Alg.~\ref{alg:train_theta}.
    The selection process begins by training a surrogate model on $\mathcal{D}^\text{A}$. All previous propositions assume an optimal model $\vtheta^{A^*}$ is being utilized.
\end{minipage}


However, in real-world scenarios, such a perfect model is unattainable. To address this, we prove that the worst-case surrogate model parameters remain within a close neighborhood of the optimal model (Supp.~Sec.~\ref{supp-ssec:minimax}). This justifies the replacement of optimal models with surrogate models, $\hat\vtheta^{\text{bat}|\text{A}}_{\text{surrogate}}$. 
Then, each backbone data point is assigned a score (Eq.~\ref{eq:score_function}) based on the surrogate model.
Data points with scores exceeding the threshold $\eta$ are selected as BAT data (i.e., $\mathcal{D}^{\text{B}^\prime}$ in Proposition~\ref{prop:p2}). Finally, the adapter is trained using the selected backbone data combined with $\mathcal{D}^\text{A}$. Examples of \batalg{} selection are shown in Supp.~Sec.~\ref{fig:qualitative_score1}.

\textbf{Normalized Loss Sampling Data Selection.}
Additionally, we propose a method called \textit{normalized loss sampling data selection} to improve the accuracy of biased scheme results for nondeterministic models. For instance, when a diffusion model generates an image, it iteratively adds and removes noise $\epsilon$, sampled from $\mathcal{N}(0, \mI)$, where $\dim(\mI)$ corresponds to the dimension of the image's latent representation. To mitigate the model's nondeterministic behavior, we simply sample the noise $\delta$ times. Unless specified, we fix $\delta=3$ in all subsequent experiments.


\section{Experiments}
\label{sec:experiment}

We have proposed a practically applicable Backbone Augmented Training (BAT) algorithm (Alg.~\ref{alg:train_theta}) in Sec.~\ref{sec:scheme}, which is mathematically validated in Sec.~\ref{ssec:assumptions}.
Beyond the theoretical validations, in this section, we now demonstrate the effectiveness BAT through experimental results.
First, we empirically validate BAT (Sec.~\ref{ssec:validation}) and demonstrate its robustness across various benchmarks (Sec.~\ref{ssec:benchmark_test}). Next, we analyze performance variations with an imperfect surrogate model (Sec.~\ref{sec:weak_surrogate}) and ablate the backbone augmentation ratio $\gamma$ (Sec.~\ref{ssec:gamma_ablation}). Finally, we show that augmenting adaptation data with backbone-like data remains effective even when direct access to $\mathcal{D}^\text{B}$ is limited (Sec.~\ref{ssec:inaccessible}).


\subsection{Empirically Validating BAT}
\label{ssec:validation}

\textbf{Settings.}
We evaluated our approach on personalization tasks using DreamBooth \citep{dreambooth} and LoCon from LyCORIS \citep{lycoris}.
For backbone data, we sampled 2,981 images from LAION \citep{laion} dataset. And for adaptation data, we utilized Star Wars dataset \citep{starwars}. 
See Supp.~Sec.~\ref{setting} for more details.

To validate BAT, we used the sample ratio, which represents the proportion of the original backbone dataset used to sample additional backbone data. 
For example, if $\gamma=0.95$, the sample ratio is 0.5, and the adaptation dataset contains 57 images, we first sample 50\% of the images from the backbone dataset and then select 3 images for $\mathcal{D}^\text{bat}$.
A lower sample ratio corresponds to a weaker application of our theoretical framework, while a higher sample ratio enforces it more strongly. As the sample ratio increases, we expect BAT to achieve better performance.



\textbf{Results.} 
Fig.~\ref{fig:result1} demonstrates that \batalg{} selects appropriate backbone data, leading to improved image quality, text fidelity, and personalization accuracy compared to random augmentation and standard adaptation.
Moreover, the proportional relationship between the sample ratio and benchmark scores empirically validates BAT.
This suggests that as BAT becomes more generalized, it aligns more closely with universal data selection principles.


\subsection{Evaluating BAT on Benchmarks}
\label{ssec:benchmark_test}

\begin{wraptable}{r}{0.5\textwidth}
    \vspace{-20pt}
    \caption{
    \textbf{BAT in Image and Language Domains.} The table shows the results of applying our BAT to DreamBooth \cite{dreambooth} and LoCon \cite{lycoris} for personalized image generation on DreamBooth Corgi dataset, while the bottom table presents the results of applying BAT acoross various benchmarks for language models (HS: HellaSwag; WG: WinoGrande).}
    \label{tab:tab1}
    \begin{center}
    \begin{small}
    \begin{sc}
    \resizebox{\linewidth}{!}{
        \begin{tabular}{@{}lcccc@{}}
        \toprule
        & Cosine Sim \(\uparrow\) & Centroid Distance \(\downarrow\)& CLIP \(\uparrow\) & Vendi \(\downarrow\) \\ 
        \midrule
        \midrule
        DreamBooth & 0.3861 & 797.8 & 0.2666 & 4.812 \\
        + BAT & \textbf{0.4189} & \textbf{695.7} & \textbf{0.3157} & \textbf{2.191} \\
        \midrule
        LoCon & 0.5427 & 824.8 & 0.4884 & 1.847 \\
        + BAT & \textbf{0.5502} & \textbf{823.5} & \textbf{0.4952} & \textbf{1.839} \\
        \bottomrule
        \end{tabular}
    }
    
    \vspace{5pt}
    
    \resizebox{\linewidth}{!}{
        \begin{tabular}{@{}lcccccccc@{}}
        \toprule
        & BoolQ & PIQA & SIQA & HS & WG & ARC-c & ARC-e & OBQA \\ 
        \midrule
        \midrule
        LoRA & 62.17 & 76.28 & 74.51 & 76.52 & 48.86 & 48.70 & \textbf{74.07} & \textbf{32.70} \\
        + BAT & \textbf{65.17} & \textbf{80.25} & \textbf{77.02} & \textbf{76.87} & \textbf{51.38} & \textbf{53.20} & 71.93 & 42.83 \\
        \midrule
            DoRA        & 62.17               & 76.50               & 72.36               & \textbf{76.42}                   & 50.28              & 37.54              & \textbf{74.96}              & \textbf{60.80}                    \\
            + BAT       & \textbf{63.96}      & \textbf{78.84}      & \textbf{74.36}      & 76.39          & \textbf{73.88}     & \textbf{42.66}     & 71.89     & 57.00           \\
        \bottomrule
        \end{tabular}
        
    }
    \end{sc}
    \end{small}
    \end{center}
    \vspace{-20pt}
\end{wraptable}

\textbf{Settings.} 
We now demonstrate that BAT outperforms standard adaptations across most benchmarks. For DreamBooth and LoCon, we used the DreamBooth Corgi dataset as the adaptation dataset, with the backbone dataset setting identical to Sec.~\ref{ssec:validation}. Additionally, we evaluated the commonsense reasoning ability of our approach on LLaMA 2-7B and 3-8B \citep{llama2} with LoRA\citep{lora} and DoRA~\citep{dora} adaptations.
Since language models typically prioritize their pre-training datasets (i.e., $ \mathcal{D}^\text{B}$), we used a publicly available version further fine-tuned on the Stanford Alpaca dataset \citep{alpaca}. In this setting, we used the Stanford Alpaca dataset as the backbone dataset and the training set of each commonsense reasoning benchmark as the adaptation dataset.

To simulate the challenges posed by limited adaptation data, we intentionally reduce both the amount of adaptation data and the number of training steps.
For DreamBooth and LoCon, we did not impose such restrictions and follow the same training setting (we showed further restricted comparisons in Supp.~Sec.~\ref{fig:qualitative}), as adaptation data are already limited in personalization tasks.
For language adaptation, we reduced the training steps by one-third, except for the HellaSwag benchmark \citep{hellaswag}, which is nearly as large as the fine-tuning dataset. In this case, we limit adaptation training to only 10\% of the total data. Additional training details are provided in Supp.~Sec.~\ref{setting}.

\textbf{Results.} Table~\ref{tab:tab1} presents the performance of \batalg{} across various benchmarks when the sample ratio is set to 1.
Most benchmarks report improved scores, except for HellaSwag (DoRA), ARC-e, and OBQA in language adaptation.
This can be attributed to two factors: (1) some benchmarks are highly dependent on the downstream task, limiting the amount of backbone data that can effectively enhance adaptation training; (2) the adaptation dataset may already contain a sufficient amount of data.
Nonetheless, in most cases, BAT achieved superior benchmark scores.

\subsection{Strong Versus Weak Surrogates}
\label{sec:weak_surrogate}

\begin{wraptable}{r}{0.5\textwidth}
    \vspace{-20pt}
    \caption{
    \textbf{Strong vs. for Weak Surrogate.} This table shows DreamBooth LoRA adaptation benchmark results on Kaggle Star Wars dataset \citep{starwars} with BAT using different surrogate models. 
    }
    \label{tab:strong_vs_weak}
    \begin{center}
    \begin{small}
    \begin{sc}
    \resizebox{\linewidth}{!}{
        \begin{tabular}{@{}lcccc@{}}
        \toprule
        & Cosine & Centroid & \multirow{2}{*}{CLIP \(\uparrow\)} & \multirow{2}{*}{Vendi \(\downarrow\)} \\ 
        & Sim \(\uparrow\)& Distance \(\downarrow\)& & \\ 
        \midrule
        \midrule
        DreamBooth (baseline) & 0.4106 & 628.3 & 0.2570 & 8.461 \\
        \midrule
        Weak (200 steps) & 0.4504 & 585.9 & 0.2639 & 7.616 \\
        Weak (400 steps) & 0.4538 & 596.1 & 0.2688 & 7.919 \\
        \midrule
        Strong (800 steps) & 0.4778 & 562.8 & 0.2792 & 6.803 \\
        \bottomrule
        \end{tabular}
    }
    \end{sc}
    \end{small}
    \end{center}
    \vspace{-20pt}
\end{wraptable}

\textbf{Settings.} Training surrogate models can be demanding. So, in this section, we investigate the effect of reducing the training overhead in surrogate training. We created weaker surrogates by reducing the training steps to half or even to quarter compared to the original surrogate.
All the other settings are identical to Sec.~\ref{ssec:benchmark_test} except adaptation data being Kaggle Star Wars dataset.





\textbf{Results.} Table~\ref{tab:strong_vs_weak} shows that weaker surrogate models outperform the original adaptation in benchmark evaluations. This demonstrates that reducing training steps of the surrogate model can still improve adaptation performance, thereby reducing the computational demand of applying BAT.



\subsection{Backbone Augmentation Ratio Ablation}
\label{ssec:gamma_ablation}

\textbf{Settings.}
In this experiment, we analyze the impact of varying the backbone augmentation ratio $\gamma$ (Def.~\ref{def:def2}). This ratio determines the amount of backbone data in $\mathcal{D}^{\text{bat}|\text{A}}$.
Starting from $\gamma=0.5$ (i.e., the added backbone data equals the adaptation data), we gradually increase it and evaluate the model's performance on DreamBooth benchmarks.

\begin{figure}
    \vspace{-5pt}
    \begin{center}
    \includegraphics[width=1\columnwidth]{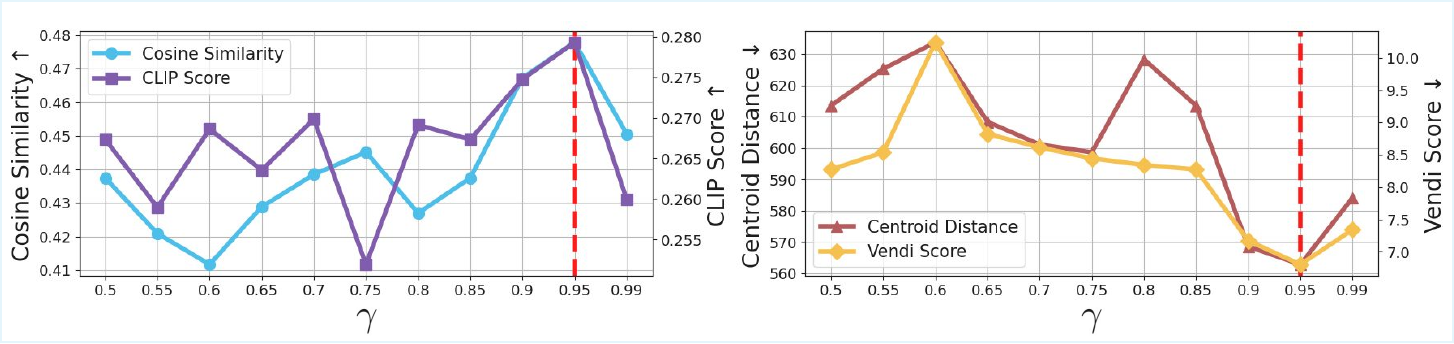}
    \vspace{-20pt}
    \caption{
    \textbf{Results of Backbone Augmentation Ratio Test.} This figure displays the benchmark scores on different backbone augmentation ratio $\gamma$ with identical adaptation dataset. This simply means more adaptation data is replaced with backbone data with BAT as $\gamma$ increases.}
    \label{fig:gamma}
    \end{center}
    \vspace{-10pt}
\end{figure}

\textbf{Results.} Fig.~\ref{fig:gamma} shows that at a certain augmentation ratio ($\gamma=0.95$), BAT achieved peak performance across all metrics.
According to theoretical and empirical studies on biased selection schemes with toy models, \cite{keypaper} demonstrated that a toy model's performance is maximized when the proportion of selected and original data reaches an optimal point. Until now, biased schemes were not extended to current large models.
Through this experiment, we confirm that our propositions and the \batalg{} algorithm, which incorporates a biased selection scheme in BAT, align with the data selection theory.


\subsection{Inaccessible Backbone Data} 
\label{ssec:inaccessible}
\begin{wraptable}{r}{0.6\textwidth}
    \vspace{-18pt}
    \caption{
    \textbf{Effect of Our Data Augmentation Using External Data.} 
    Assuming backbone data is inaccessible, performance is evaluated using our data selection algorithm on data outside the backbone.
    }
    \label{tab:tab2}
    \begin{center}
    \begin{small}
    \begin{sc}
    \resizebox{\linewidth}{!}{
        \begin{tabular}{@{}lcccc@{}}
        \toprule
        & Cosine & Centroid & \multirow{2}{*}{CLIP \(\uparrow\)} & \multirow{2}{*}{Vendi \(\downarrow\)} \\ 
        & Sim \(\uparrow\)& Distance \(\downarrow\)& & \\ 
        \midrule
        \midrule
        DreamBooth & \textbf{0.3861} & 797.8 & 0.2666 & 4.812 \\
        + BAT w/o backbone data & 0.3659 & \textbf{795.8} & \textbf{0.2919} & \textbf{4.722} \\
        \bottomrule
        \end{tabular}
    }
    \end{sc}
    \end{small}
    \end{center}
    \vspace{-20pt}
\end{wraptable}

\vspace{-5pt}

\textbf{Settings.} Some large models do not publicly disclose their pre-training data $\mathcal{D}^\text{B}$ \citep{gpt3, sd3}. However, their input and output features remain accessible. Leveraging this feature information, we can select open-source data with similar distributional characteristics at both individual data point and dataset levels. While this study does not introduce theoretical modifications for this scenario, we explore its implications by applying alternative datasets that are not part of the original backbone dataset.
Specifically, we collected 50 images from Google Images using the query ``Corgi Dog" that closely resemble the DreamBooth Corgi dataset. We then applied \batalg{}, setting $\gamma = 0.95$, with this online dataset as the backbone dataset.

\textbf{Results.} Table~\ref{tab:tab2} shows that while this approach performs worse than BAT when adaptation data are accessible, it still outperforms adaptations without BAT. This result seems counterintuitive as Google-selected images should yield better performance than the original BAT since they appear more similar to the adaptation data from a human perspective.
This is because, unlike human perception, Google-selected images may be farther from the distribution of both the backbone model and the adaptation data, diminishing performance. More ablation can be found in Supp. Sec.~\ref{more_quan}

\section{Conclusion}
\label{sec:conc}

\vspace{-5pt}

This study addresses the challenges posed by limited adaptation data through the introduction of Backbone Augmented Training (BAT). First, we present two well-founded propositions that provide the theoretical rationale (Proposition~\ref{prop:p1}) and necessary conditions (Proposition~\ref{prop:p2}) for BAT, supported by rigorous proofs. Building on these mathematical foundations, we propose an efficient BAT algorithm, \batalg{} (Alg.~\ref{alg:train_theta}), designed to operate within practical time complexity. Finally, we validate the effectiveness of BAT through experimental results, demonstrating its capability to enhance adaptation performance.

\textbf{Limitations and Future Works.} For Sec.~\ref{ssec:inaccessible}, our study only presents the empirical results of the alternative of backbone data. Though it is out of our study's scope, more rigorous mathematical investigation is required in utilizing the data outside of adaptation and backbone data. This will open broader applications for data augmentation for adaptations.

\newpage

\bibliography{bat}
\bibliographystyle{unsrt}

\newpage
\appendix
\onecolumn


\section{Mathematical Supplements}
\label{supp-sec:math-supple}


\subsection{Basic Notations}
\label{ssec:notation}

For standard notations, we denote consistency of random variables as \( \textstyle X_n \overset{P}{\rightarrow} X \). Next, using the notation $\displaystyle {p\!\!-\!\!\lim}$, we define probabilistic asymptotics as
\begin{equation} 
    X_n = o_{P}(a_n) \iff \underset{n \rightarrow \infty}{p\!\!-\!\!\lim}\frac{|X_n|}{a_n} = 0,
\end{equation}
and denote almost sure convergence as
\begin{equation} 
    X_n \xrightarrow{a.s.} X \iff \underset{n \rightarrow \infty}{\lim}P(X_n = X) = 1.
\end{equation}
Lastly, for some matrices $\displaystyle \mX \text{ and } \mY$, we denote $\displaystyle \mX \succeq \mY$ if $\displaystyle \mX - \mY$ is positive semi-definite, and $\displaystyle \mX \succ \mY$ if it is positive definite.

Now, for a parameter space $\Theta$ and an estimator $\displaystyle \vtheta_n \in \Theta$, we define an empirical risk function $R_n: \Theta \rightarrow \R$ as
\begin{equation} 
    R_n(\vtheta_n) := \frac{1}{n}\sum_{i = 1}^n\mathcal{L}_{i}^{\vtheta} \iff R^{\vtheta}_n = R_n(\vtheta_n),
\end{equation}
where $\displaystyle \mathcal{L}_{i}^{\vtheta} := \mathcal{L}(Y_i, f(X_i;\vtheta_i)).$
Here, $\displaystyle \mathcal{L}: \R^{d} \times \R^{d} \rightarrow \R$ represents the loss function of the parameters, where $\displaystyle f(\cdotp;\vtheta): \R^{p} \rightarrow \R^{d}$ refers to the model.
Additionally, $X_i$ and $Y_i$ represent the sampled input and label at training step $i$.
In the case of deterministic sampling, these are denoted using lowercase notations $\vx$ and $\vy$.

Now, by the law of large numbers, we can define some $\displaystyle R$ satisfying $\displaystyle R_n \xrightarrow{P} R$. We set $\displaystyle \hat\vtheta_n$ to be the nearly minimizing estimator that satisfies the following condition:
\begin{equation} 
    R_n(\hat\vtheta_n) \le \inf_{\vtheta \in \Theta} R_n(\vtheta_n) + o_P(1).
\end{equation}

Recall that every risk in this study uses sampled sets to optimize their corresponding models. To discuss convergence across the entire sample, we must define the total risk. We can achieve this with a simple expectation as follows:
\begin{equation} 
    R(\vtheta) := \E\mathcal{L}(\vy, f(\vx;\vtheta)),
\end{equation}
where $(\vx,\vy)$ are independent and identically distributed (i.i.d.) according to some distribution $P(\cdot)$.

\subsection{Assumptions}
\label{assumptions}

In this section, we introduce textbook assumptions for asymptotic probability. These are just applied versions of regularity conditions.

\begin{enumerate}[itemsep=0pt]
    \item[\textit{A1.}] \label{asm:assumption1}
    $\displaystyle R^\text{B}$ \textit{ and } $R^\text{A}$ \textit{are minimized uniquely at} $\displaystyle \vtheta^{\text{B}^*}$ \textit{and} $\vtheta^{\text{A}^*}$, \textit{respectively.}
    \item[\textit{A2.}] $\displaystyle \mathcal{L}^\text{B}$ \textit{and} $\mathcal{L}^\text{A}$ \textit{are both greater than zero and always lower semi-continuous. Moreover, for every} $\displaystyle \vu \in \mS^{dim(\Theta^\text{B}) - 1}$ \textit{and} $g(\vu) \in \mS^{dim(\Theta^\text{A}) - 1}$, \textit{define} $\displaystyle \mathcal{L}_{\infty}^\text{B}$ \textit{and} $\mathcal{L}_{\infty}^\text{A}$ \textit{both in} $\overline \R_{\ge0}$ \textit{as follows:}
    \begin{equation} 
        \begin{cases}
        \mathcal{L}_{\infty}^\text{B}(\vu; \vx, \vy) := \underset{\underset{\vtheta/ ||\vtheta|| \rightarrow \vu}{||\vtheta|| \rightarrow \infty}}{\liminf}\mathcal{L}^\text{B}, \\
        \mathcal{L}_{\infty}^\text{A}(g(\vu); \vx, \vy) := \underset{\underset{\vtheta/ ||\vtheta|| \rightarrow g(\vu)}{||\vtheta|| \rightarrow \infty}}{\liminf}\mathcal{L}^\text{A}.
        \end{cases}
        \nonumber
    \end{equation}
    \textit{Then, the following holds:}
    \begin{equation}  \inf_\vu\E\mathcal{L}_{\infty}^\text{B} > R(\vtheta^{\text{B}^*}) \text{ and } \inf_{g(\vu)}\E\mathcal{L}_{\infty}^\text{A} > R(\vtheta^{\text{A}^*}).
    \nonumber
    \end{equation}
    \item[\textit{A3.}] \textit{Both} $\displaystyle \mathcal{L}^{\vtheta^\text{B}}$ \textit{ and } $\mathcal{L}^{\vtheta^\text{A}}$ are differentiable at $\displaystyle \vtheta^{\text{B}^*}$ \textit{ and } $\vtheta^{\text{A}^*}$ \textit{respectively, for} $\displaystyle \mathbb{P}$\textit{-almost all} $\displaystyle (\vy, \vx)$. \textit{Further, for a neighborhood} $\displaystyle U$ \textit{of} $\displaystyle \vtheta^{\text{B}^*}$ \textit{ or } $\vtheta^{\text{A}^*}$, \textit{the following holds:}
    \begin{equation} 
        \E\sup_{\vtheta_1 \neq \vtheta_2 \in U}\left[\frac{|\mathcal{L}(\vtheta_1)-\mathcal{L}(\vtheta_2)|}{||\vtheta_1-\vtheta_2||^2_2}\right] < \infty.
        \nonumber
    \end{equation}
    \item[\textit{A4.}] $\displaystyle R^\text{B}$ \textit{and} $R^\text{A} \in C^2$ \textit{exists, with} $\displaystyle \mH^\text{B}(\vx), \mH^\text{A}(\vx) \succeq \mathbf{0}.$ 
\end{enumerate}

\subsection{Lemma 1} \label{supp-ssec:lemma1} 
\textit{Assume that the map $\displaystyle \mathcal{L}^{\vtheta}(\vx): \Theta \rightarrow \R$ is lower semi-continuous for almost all $\displaystyle \vx$ which is any input data of the estimator. Then, for any $\displaystyle \vtheta \in \Theta$},
\begin{equation}
    \mathcal{L}^{\vtheta}(\vx) \le \underset{\vtheta_n \rightarrow \vtheta}{\liminf}\mathcal{L}^{\vtheta_n}(\vx), \quad \textit{almost surely.}
\end{equation}

\textbf{Proof of Lemma 1.}
We begin by recalling the definition of lower semi-continuity. A function \( f: \Theta \to \mathbb{R} \) is lower semi-continuous at \( \theta \) if:
\begin{equation}
    \liminf_{\theta_n \to \theta} f(\theta_n) \geq f(\theta).
\end{equation}
This property ensures that the function does not suddenly drop in value near \( \theta \). Formally, for any sequence \( \theta_n \to \theta \), we have:
\begin{equation}
\liminf_{n \to \infty} f(\theta_n) \geq f(\theta).
\end{equation}
Given that \( \mathcal{L}^\theta(x) \) is lower semi-continuous for almost all \( x \), we can apply the definition of lower semi-continuity. Specifically, for any \( \theta \in \Theta \) and any sequence \( \theta_n \to \theta \), it follows that:
\begin{equation}
\mathcal{L}^\theta(x) \leq \liminf_{\theta_n \to \theta} \mathcal{L}^{\theta_n}(x).
\end{equation}
This inequality holds because \( \mathcal{L}^\theta(x) \) is assumed to be lower semi-continuous.

The term \textit{almost surely} means that the inequality holds for almost all values of \( x \). In other words, there may be a set of measure zero where the inequality does not hold, but this set is negligible.

Thus, for almost every \( x \) (except on a set of measure zero), the following inequality holds:
\begin{equation}
\mathcal{L}^\theta(x) \leq \liminf_{\theta_n \to \theta} \mathcal{L}^{\theta_n}(x), \quad \text{almost surely.}
\end{equation}
By combining these observations, we conclude that since \( \mathcal{L}^\theta(x) \) is lower semi-continuous for almost all \( x \), for any sequence \( \theta_n \to \theta \), the lemma is proven. \hfill \(\Box\)

\subsection{Lemma 2} \label{supp-ssec:lemma2} 
\textit{For any sufficiently small neighborhood $\displaystyle U \subset \Theta$ around $\displaystyle \vtheta$, if the map $\displaystyle {\inf}_{\vtheta \in U}\mathcal{L}^{\vtheta}(\vx): \R^\vp \rightarrow \R$ satisfies the condition of Theorem 1, then the map is measurable and $\displaystyle R(\vtheta) > -\infty$ for $\displaystyle \vtheta$ that satisfies $\displaystyle {\inf}_{\vtheta \in U}\mathcal{L}^{\vtheta}$.}

\textbf{Proof of Lemma 2.} Using Lemma 1 (Sec.~\ref{supp-ssec:lemma1}), we know that if \( \mathcal{L}^\theta(x) \) is lower semi-continuous, then for any \( \theta \in \Theta \):
\begin{equation}
\mathcal{L}^\theta(x) \leq \liminf_{\theta_n \to \theta} \mathcal{L}^{\theta_n}(x) \quad \text{almost surely}.
\end{equation}

Analyzing the map \( \inf_{\theta \in U} \mathcal{L}^\theta(x) \), which is the infimum of \( \mathcal{L}^\theta(x) \) over a neighborhood \( U \subset \Theta \) around \( \theta \), the function \( \mathcal{L}^\theta(x) \) is assumed to satisfy the lower semi-continuity condition of Lemma 1 (Sec.~\ref{supp-ssec:lemma1}).

We now show that the map \( \inf_{\theta \in U} \mathcal{L}^\theta(x) \) is measurable. Since lower semi-continuous functions are measurable in standard measure theory, we conclude that \( \mathcal{L}^\theta(x) \) is measurable. Further, the infimum of a collection of lower semi-continuous functions over a compact set is itself lower semi-continuous, and hence measurable.

Next, define \( R(\theta) = \inf_{\theta \in U} \mathcal{L}^\theta(x) \). We need to show that \( R(\theta) > -\infty \). Since \( \mathcal{L}^\theta(x) \in \mathbb{R} \) is bounded from below and lower semi-continuous on a compact set, the infimum will also be bounded from below. Hence, \( R(\theta) > -\infty \).

Thus, this lemma is proven. \hfill \(\Box\)

\subsection{Lemma 3}
\label{supp-ssec:lemma3} 
\textit{Let the map $\displaystyle \mathcal{L}^{\vtheta}(\vx): \Theta \rightarrow \R$ satisfies the conditions for Lemma 1 (Sec.~\ref{supp-ssec:lemma1}) and 2 (Sec.~\ref{supp-ssec:lemma2}). Then, for any nearly minimizing estimator $\displaystyle \hat\vtheta_n$ and some globally minimizing parameter $\displaystyle \vtheta^* \in \Theta^*$ for some global minimum space in case there are multiple or continuous set of globally minimizing parameters, for any $\displaystyle \varepsilon > 0$ and compact set $\displaystyle A \subset \Theta$,} 
\begin{equation} 
    P(\text{dist}(\hat\vtheta_n, \Theta^*) \ge \varepsilon \land \hat\vtheta_n \in A) \rightarrow 0.
\end{equation}

\textbf{Proof of Lemma 3.}

\textit{Case 1.} For all $\displaystyle \vtheta \in \Theta$, assume $\displaystyle R(\vtheta) = \infty$, then by the assumption of nearly minimum and derivation with the law of large number like above, $R_n(\hat\vtheta_n) \le R(\vtheta^*) + o_P(1)$.
This makes all $R_n(\hat\vtheta_n)$ converge to $\displaystyle \infty$ in probability, letting $\displaystyle \Theta = \Theta^*$ and $\displaystyle \text{dist}(\hat\vtheta_n, \Theta^*) \xrightarrow{P} 0$. 
Now, for the case where for some $\displaystyle \vtheta^*$ such that $\displaystyle R(\vtheta^*) < \infty$, let $\displaystyle U_{m}{\downarrow}\vtheta$ be a diminishing sequence of open neighborhoods around a chosen $\displaystyle \vtheta$ as their diameters converge to zero. Then, by the assumption of Lemma 2 (Sec.~\ref{supp-ssec:lemma2}), $\displaystyle R(\vtheta^*) > -\infty$ when $\displaystyle \mathcal{L}^{\vtheta^*} = |\mathcal{L}^{\vtheta^*}|$ for all $\displaystyle X \text{ and } Y.$

Denote $\displaystyle \mathcal{L}^U(\vx)$ for $\displaystyle {\inf}_{\vtheta \in U}\mathcal{L}^\vtheta(\vx)$. The sequence $\displaystyle \mathcal{L}^{U_m}$ is increasing and lower than $\displaystyle \mathcal{L}^\vtheta$ by its definition. Then, by Lemma 1 (Sec.~\ref{supp-ssec:lemma1}), regarding $\displaystyle \vtheta_n \rightarrow \vtheta$, as some $\displaystyle \vtheta' \in {U_m} \rightarrow \vtheta$, $\displaystyle \mathcal{L}^{U_m}$ is the left-hand limit of $\displaystyle \mathcal{L}^\vtheta$ almost surely. Recall the monotone convergence theorem \citep{measure_theory}, then by the definition of $\displaystyle R$ which involves expectation and integral, $\displaystyle {R^U}(\vtheta_m)$ where $\displaystyle \vtheta_i \text{ satisfies } \mathcal{L}^{U_i}$ is also the left-hand limit of $\displaystyle R(\vtheta)$.

\textit{Case 2.} For $\displaystyle \vtheta \notin \Theta^*$, $\displaystyle R(\vtheta) > R(\vtheta^*)$ by definitions. Then, from the proceeded arguments, there exists an open neighborhood $\displaystyle U^\vtheta$ of $\displaystyle \vtheta$ where $\displaystyle R(\vtheta) > R(\vtheta^*)$. This implies that the set $\displaystyle B = \{\vtheta \in A: dist(\vtheta, \Theta^*) \ge \varepsilon\}$ is compact as it is covered by the subset of $\displaystyle \{U^\vtheta: \vtheta \in B\}$.

Let $\displaystyle U^{\vtheta_1}, U^{\vtheta_2}, \dots, U^{\vtheta_p}$ be such subcovers. By the law of large numbers and definition of $\displaystyle U$, 
\begin{equation} 
    \underset{j = 1,\dots, p}{\inf}R_n^U(\vtheta_j) \le \underset{\vtheta \in B}{\inf}R_n(\vtheta) \xrightarrow{a.s.} R(\vtheta^*) < \underset{j}{\inf}R^U(\vtheta_j).
\end{equation}
If $\displaystyle \hat\vtheta_n \in B$, then $\displaystyle {\inf}_{\vtheta \in B}R_n(\vtheta)$ is less than or equal to $\displaystyle R_n(\hat\vtheta)$ by $\displaystyle B$'s definition. Then by the definition of $\displaystyle \hat\vtheta_n$, $\displaystyle {\inf}_{\vtheta \in B}R_n(\vtheta)$ is also less than or equal to $\displaystyle R_n(\vtheta^*)$ and also less than or equal to $\displaystyle R(\vtheta^*)$ as $\displaystyle n \rightarrow \infty$ by the consistency of $\displaystyle R_n$ covered under the definition of it. So,
\begin{equation}
    \{\hat\vtheta \textit{ } | \textit{ } \hat\vtheta \in B\} \subset \{\underset{\vtheta \in B}{\inf}R_n(\vtheta) \le R(\vtheta^*) + o_P(1)\}.
\end{equation}
This means that the probability of the event on the right side, which is the equivalent to the last line of the lemma, converges to zero, proving this lemma. \hfill $\displaystyle \square$

\subsection{Proof of Proposition 1.} \label{supp-ssec:p1_proof} 
For $\displaystyle ||\boldsymbol\zeta|| < 1$, define $\displaystyle \boldsymbol\varphi(\boldsymbol\zeta) = r(||\boldsymbol\zeta||)\boldsymbol\zeta$ with $\displaystyle r(c) = 1/(1-c^2)$ to deal with more concentrated parameters than unit parameters, then define the loss for backbone augmented adaptation,
\begin{equation} 
        \mathcal{L}^{\text{bat}|\text{A}}(\boldsymbol\zeta; \vx, \vy) := 
    \left\{
    \begin{array}{@{}l@{\thinspace}l}
        \mathcal{L}^{\text{bat}|\text{A}}(\boldsymbol\varphi(\boldsymbol\zeta); \vx, \vy) & \mbox{if } ||\boldsymbol\zeta|| < 1, \\
        \mathcal{L}^{\text{bat}|\text{A}}_{\infty}(\boldsymbol\zeta; \vx, \vy) &  \mbox{if } ||\boldsymbol\zeta|| = 1,
    \end{array}
    \right.
\end{equation}
so that 
\begin{equation}
    R^{\text{bat}|\text{A}}(\boldsymbol\zeta) = \E\mathcal{L}^{\text{bat}|\text{A}}(\boldsymbol\zeta, \vx, \vy), \quad
    R^{\text{bat}|\text{A}}_k = k^{-1}\textstyle\sum^k_i\mathcal{L}^{\text{bat}|\text{A}}(\boldsymbol\zeta, \vx_i, \vy_i),
\end{equation} 
for $\displaystyle \boldsymbol\zeta \in \mathbb{B}^{dim(\Theta^\text{A})}(1)$ which is a unit ball in $\displaystyle \Theta^\text{A}$ and $\displaystyle (\vx, \vy) \in G.$ Suppose that 
\begin{equation}
    \hat{\boldsymbol{\zeta}} := \argmin_{\boldsymbol{\zeta} \in \mathbb{B}^{dim(\Theta^\text{A})}}(|R^{\text{A}}_{N^{A}}(\vtheta^{\text{A}^*}) -R^{\text{bat}|\text{A}}_k(\hat\vtheta^{\text{bat}|\text{A}}_k)| - |R^\text{A}_{N^{A}}(\vtheta^{\text{A}^*}) - R^{\text{bat}|\text{A}}_k(\vtheta^{\text{A}^*})|), \\
\end{equation}
\begin{equation}
    \boldsymbol{\zeta}^* := \argmin_{\boldsymbol{\zeta} \in \mathbb{B}^{dim(\Theta^\text{A})}}(|R^{\text{A}}(\vtheta^{\text{A}^*}) - R^{\text{bat}|\text{A}}(\hat\vtheta^{\text{bat}|\text{A}})| - |R^\text{A}(\vtheta^{\text{A}^*}) - R^{\text{bat}|\text{A}}(\vtheta^{\text{A}^*})|).
\end{equation}

The second term is unique from Assumption 5. Recall that $\displaystyle \mathcal{L}^{\text{bat}|\text{A}}$ is defined on both $\displaystyle \mathcal{D}^\text{B} \text{ and } \mathcal{D}^\text{A}.$ We know that $\displaystyle \mathcal{L}^{\text{bat}|\text{A}}$ defined on $\displaystyle \mathcal{D}^\text{A}$ is simply $\mathcal{L}^\text{A}$ as $\vtheta^{\text{bat}|\text{A}} \in \Theta^\text{A}$ by definition. Thus, the continuity feature is demonstrated. However, for $\displaystyle \mathcal{L}^{\text{bat}|\text{A}}$ defined on $\displaystyle \mathcal{D}^\text{B}$, one has to use the nature of adaptation to depict the lower semi-continuity. 

Since $\displaystyle \mathcal{L}$ is a compositional function of $\displaystyle f$, $\displaystyle \vtheta$, and $\displaystyle (\vx, \vy)$, showing $\displaystyle f$'s lower semi-continuity will be enough. Then, we want to show that $\displaystyle f^\text{A}(\vx_{_\text{B}};\vtheta^\text{A})$ has lower semi-continuity when $\displaystyle (\vx_{_\text{B}}, \vy_{_\text{B}}) \in \mathcal{D}^\text{B}.$ By the nature of adaptation regarding $\displaystyle \Delta(\vtheta^\text{A} \backslash \vtheta^\text{B})$, 
\begin{equation}
    f^\text{B}(\vx; \vtheta^{\text{B}^*}) = f^\text{A}(\vx_{_\text{B}}, \vtheta^{\text{B}^*}) - f^{\text{B} \backslash \text{A}}(\vx_{_\text{B}}, \vtheta^{\text{B}^*} \backslash \vtheta^{\text{A}}),
\end{equation}
when $\displaystyle f^{\text{B} \backslash \text{A}}$ is some function that satisfies the nature of adaptation.

Then, by Assumption 2 and the fact about the summation of lower semi-continuous functions, $\displaystyle f^\text{A}(\vx_{_\text{B}}, \vtheta^{\text{B}^*})$ is continuous. Then, by the definition of $\displaystyle g$ and nature of composition of continuous functions, $\displaystyle f^\text{A}(\vx_{_\text{B}}, g(\vtheta^{\text{B}^*})) = f^\text{A}(\vx_{_\text{B}}, \vtheta^\text{A}_1)$ also holds lower semi-continuity. Now, by Lemma~2 (Sec.~\ref{supp-ssec:lemma2}), $\displaystyle \hat{\boldsymbol{\zeta}} \rightarrow \boldsymbol{\zeta}^*$ almost surely. By Assumption 2, we get $\displaystyle ||\boldsymbol{\zeta}|| < 1$, then almost surely, $\displaystyle \hat\vtheta^{\text{bat}} \rightarrow \vtheta^{\text{A}^*} = \boldsymbol\varphi(\boldsymbol{\zeta}^*).$ Then by 
Lemma~3 (Sec.~\ref{supp-ssec:lemma3}) with Assumption~3, Assumption~4, and the argument above, the proof is completed. \hfill $\displaystyle \square$

\subsection{Proof of Proposition 2} \label{supp-ssec:p2_proof} 
By Def.~\ref{def:def2}, one can derive from the assumption,
\begin{equation}
    \frac{1}{k} ||(\mH^{\text{bat}|\text{A}})^{-1}\sum_{\mathcal{D}^\text{bat}}\nabla_\vtheta\mathcal{L}^{\text{bat}|\text{A}}|| \le \frac{1}{N^{A}}||(\mH^{\text{bat}|\text{A}} - \mH^{\text{bat}})^{-1}\sum_{\mathcal{D}^\text{A}}\nabla_\vtheta\mathcal{L}^{\text{bat}|\text{A}}|| + o_P(1),
\end{equation}
then, using the fact that $\displaystyle \mathcal{L}^{\text{bat}|\text{A}} \rightarrow \mathcal{L}^{\text{A}^*}$ by Proposition\ref{prop:p1} and the nature of adaptation regarding $\displaystyle (\vtheta^\text{A} \backslash \vtheta^\text{B})$, one can derive that $\displaystyle \mH^{\text{bat}|\text{A}} - \mH^{\text{bat}} = \mH^{\text{A}}.$ With these facts, 
\begin{equation}
        \frac{1}{k} ||(\mH^{\text{bat}|\text{A}})^{-1}\sum_{\mathcal{D}^\text{bat}}\nabla_\vtheta\mathcal{L}^{\text{bat}|\text{A}}|| \le \frac{1}{N^{A}}||(\mH^{\text{A}})^{-1}\sum_{\mathcal{D}^\text{A}}\nabla_\vtheta\mathcal{L}^{\text{A}}|| + o_P(1).
\end{equation}
is given. Then, by a using Newton's method, we can define,
\begin{equation}
        \hat\vtheta_k^{\text{bat}|\text{A}} - \vtheta^{\text{A}^*} = \frac{1}{k}(\mH^{\text{bat}|\text{A}})^{-1}\sum_{\mathcal{D}^\text{bat}}\nabla_\vtheta\mathcal{L}^{\text{bat}|\text{A}},
\end{equation}
\begin{equation}
    \hat\vtheta_n^\text{A} - \vtheta^{\text{A}^*} = \frac{1}{{N^{A}}}(\mH^{\text{A}})^{-1}\sum_{\mathcal{D}^\text{A}}\nabla_\vtheta\mathcal{L}^{\text{A}},
\end{equation}
and with this, we can show that $\rho$ is
\begin{equation}
\label{rho}
    \E\text{Tr}(\nabla_\vtheta\mathcal{L}{\nabla_\vtheta\mathcal{L}}^T\mH^{-1}\mS\mH^{-1}),
\end{equation}
and by combining the facts above, the proposition is proven. Also, recall that $\displaystyle \gamma \rightarrow 1$ will cause $\displaystyle \mH^{\text{bat}} \rightarrow \mathbf{0} \text{ and } \textstyle\sum_{\mathcal{D}^{\text{B}^\prime}}\nabla_{\vtheta}\mathcal{L}^{\text{bat}|\text{A}} \rightarrow 0$ by definitions proving the last part of the argument. \hfill $\displaystyle \square$

\subsection{Proposition 1 for Specific Adaptations}
\label{supp-ssec:proof_adapters}

\textbf{Proposition 1 for DreamBooth.} First, the loss function of DreamBooth \cite{dreambooth} is as follows: 

\begin{align}
\mathbb{E}_{x, c, \epsilon, \epsilon', t} \Big[ w_t \|\hat{x}_{\theta}(\alpha_t x + \sigma_t \epsilon, c) - x\|_2^2 + \lambda w_t' \|\hat{x}_{\theta}(\alpha_t' x_{\text{pr}} + \sigma_t' \epsilon', c_{\text{pr}}) - x_{\text{pr}}\|_2^2 \Big]. \label{eq:db_loss}
\end{align}

$\displaystyle x$ is the latent that is going through the diffusion steps and $\displaystyle c$ is the text guidance. $\displaystyle \epsilon$ shows the noise prediction added in the latent each steps, $\displaystyle t$. Other variables are hyper-parameters to control the training ~\citep{dreambooth}.

We can easily see that DreamBooth satisfies Assumptions 2, 3, and 4 of Proposition~\ref{prop:p1} (Sec.~\ref{ssec:assumptions}) as DreamBooth and diffusion model are considered to be learnable models. Let $\displaystyle \vtheta^\text{db}$ and $\displaystyle \vtheta^\text{D}$ represent the parameters of DreamBooth and diffusion model correspondingly. Then, we observe that $\displaystyle \vtheta^\text{db}_n$ is a nearly minimizing estimator. Also, we see that
\begin{equation}
    g(\vtheta^\text{D}) = \vtheta^\text{db}_1 \Rightarrow g = \mathbf1_\mathrm{identity},
\end{equation}
as DreamBooth does not alter diffusion model parameters in the initializing step. Also, note that 
\begin{equation}
    g_2(\vtheta^\text{D}) = g(\vtheta^\text{D}) - \frac{\partial \E}{\partial \vtheta^\text{db}},
\end{equation}
for $\displaystyle \E$ is Eq.~\ref{eq:db_loss} which is shown to be continuous and by definition of partial derivation $\displaystyle g_2$ is continuous. We can use the same argument with all $\displaystyle g_n$ with $\displaystyle n > 2$. Thus, we have shown that $\displaystyle g$ is continuous, and by Proposition~\ref{prop:p1}, DreamBooth can converge faster with backbone augmentation. \hfill $\displaystyle \square$

\textbf{Proposition 1 for LoRA.} Similar to the case of DreamBooth showing LoRA \cite{lora} continuity will be sufficient to justify Backbone Augmented Training (BAT). Given a pre-trained weight matrix $\displaystyle \mW_0 \in \mathbb{R}^{d \times k}$, LoRA decomposes the weight update $\displaystyle \Delta \mW \in \mathbb{R}^{d \times k}$ into the product $\displaystyle \mB\mA$ to get the adapted matrix $\displaystyle \mW = \mW_0 + \Delta \mW$. Here, $\displaystyle \mB \in \mathbb{R}^{d \times r}$ and $\displaystyle \mA \in \mathbb{R}^{r \times k}$ with $r \ll \min\{d,k\}$. 
To prove that LoRA is continuous, we need to show that the function $\displaystyle g(\mA, \mB) = \mW_0 + \mA \mB$ is continuous.
A function $\displaystyle g: \R^{d \times r} \times \R^{r \times k} \text{ and } \R^{d \times k}$ is continuous at \( (\mA_0, \mB_0) \) if for every \( \varepsilon > 0 \), there exists a \( \delta > 0 \) such that:
\begin{equation}
\|(\mA, \mB) - \mA_0, \mB_0)\| < \delta \quad \text{implies} \quad \|f(\mA, \mB) - f(\mA_0, \mB_0)\| < \varepsilon.
\end{equation}
The function \( g(\mA, \mB) = \mW_0 + \mA \mB \) involves matrix multiplication, which is continuous. The addition of \( \mW_0 \) is constant and does not affect continuity. Hence, we need to show that the mapping \( (\mA, \mB) \mapsto \mA \mB \) is continuous. Given small perturbations \( \Delta \mA \) and \( \Delta \mB \), we have:
\begin{equation}
g(\mA + \Delta \mA, \mB + \Delta \mB) = \mW_0 + (\mA + \Delta \mA)(\mB + \Delta \mB).
\end{equation}
We expand the expression:
\begin{equation}
\mW_{\text{LoRA}} + \Delta \mW_{\text{LoRA}} = \mW_0 + \mA \mB + \mA \Delta \mB + \Delta \mA \mB + \Delta \mA \Delta \mB.
\end{equation}
The term \( \mA \Delta \mB + \Delta \mA \mB + \Delta \mA \Delta \mB \) represents the change in \( \mW_{\text{LoRA}} \) due to small perturbations in \( \mA \) and \( \mB \).

The perturbation \( \Delta \mW_{\text{LoRA}} = \mA \Delta \mB + \Delta \mA \mB + \Delta \mA \Delta \mB \) can be bounded as:
\begin{equation}
\|\Delta \mW_{\text{LoRA}}\| \leq \|\mA\| \|\Delta \mB\| + \|\Delta \mA\| \|\mB\| + \|\Delta \mA\| \|\Delta \mB\|.
\end{equation}
As \( \|\Delta \mA\| \to 0 \) and \( \|\Delta \mB\| \to 0 \), the perturbation \( \|\Delta \mW_{\text{LoRA}}\| \to 0 \). Therefore, for any \( \epsilon > 0 \), we can find a \( \delta > 0 \) such that if \( \|\Delta \mA\| < \delta \) and \( \|\Delta \mB\| < \delta \), then \( \|\Delta \mW_{\text{LoRA}}\| < \epsilon \). \hfill $\displaystyle \square$

\subsection{Derivation of Score Function}
\label{z_derivation}

Recall the general formula (Eq. \ref{rho}) for the asymptotic error coefficient \( \rho(\mS; \mQ) \). In a non-reweighting scheme with selection probability \( \pi \), we write \( \rho(\mS; \mQ) \) as \( \rho(\pi; \mQ) \), with an abuse of notation. Explicitly, we have the following expression for \( \rho(\pi; \mQ) \):

\begin{equation}
\label{result}
\rho(\pi; \mQ) = \text{Tr}\left( \left( \mathbb{E}\left[\pi(x) \mG(x)\right] \right) \mathbb{E}\left[\pi(x) \mH(x)\right] \right)^{-1} \mQ \mathbb{E}\left[\pi(x) \mH(x)\right]^{-1})
\end{equation}

Notice that this is defined for \( \mathbb{E}\left[\pi(x) \mH(x)\right] > 0 \), and we extend it to the case where \( \mathbb{E}\left[\pi(x) \mH(x)\right] \succeq 0 \) by introducing a limiting process, as shown in the following equation:

\[
\rho(\pi; \mQ) = \lim_{\lambda \to 0^+} \text{Tr}\left( \mathbb{E}\left[\pi(x) \mG(x)\right] (\lambda \mI + \mathbb{E}\left[\pi(x) \mH(x)\right]) \right)^{-1} \mQ (\lambda \mI + \mathbb{E}\left[\pi(x) \mH(x)\right])^{-1}
\]

We now aim to minimize this function over \( \pi \) subject to the convex constraints \( \mathbb{E}\pi(x) = \gamma \), and \( \pi(x) \in [0, 1] \) for all \( x \).

We claim that a minimizer \( \pi_{\text{nr}} \) always exists. To prove this, we view \( \rho(\pi; \mQ) = F(\nu) \) as a function of the probability measure \( \nu(dx) = \pi(x) \mathbb{P}(dx)/\gamma \). In other words, \( F \) is a function defined on the space of probability measures, where the Radon-Nikodym derivative with respect to \( \mathbb{P} \) is upper bounded by \( 1/\gamma \). This space of probability measures is uniformly tight. Furthermore, if \( \nu_n \) is a sequence in this space and \( \nu_n \to \nu_\infty \) weakly, it follows by the Portmanteau’s theorem that \( \nu_\infty \) also has a Radon-Nikodym derivative with respect to \( \mathbb{P} \), which is upper bounded by \( 1/\gamma \). Hence, this domain is compact by Prokhorov's theorem.

Finally, since the mappings \( \nu \mapsto \int \mG(x) \nu(dx) \) and \( \nu \mapsto \int \mH(x) \nu(dx) \) are continuous in the topology of weak convergence (because \( \mG(x) \) and \( \mH(x) \) are continuous by assumption), the function \( F(\nu) \) is lower semi-continuous. Therefore, a minimizer \( \nu_{\text{nr}}(dx) = \pi_{\text{nr}}(x) \mathbb{P}(dx)/\gamma \) exists, with \( \pi_{\text{nr}}(x) \in [0, 1] \).

Next, we consider any minimizer \( \pi_{\text{nr}} \) and any other feasible \( \pi \). Let \( \pi_t := (1 - t) \pi_{\text{nr}} + t \pi \). By assumption, \( \mH_{\pi_{\text{nr}}} > 0 \) strictly. Then we have:

\[
\rho(\pi_t; \mQ) = \rho(\pi_{\text{nr}}; \mQ) + t \int \left( \pi(x) - \pi_{\text{nr}}(x) \right) Z(x; \pi_{\text{nr}}) \mathbb{P}(dx) + o(t)
\]

This leads to the conclusion that for any feasible \( \pi \), the following condition must hold:

\begin{equation}
\label{contra}
J(\pi; \pi_{\text{nr}}) := \int \left( \pi(x) - \pi_{\text{nr}}(x) \right) Z(x; \pi_{\text{nr}}) \mathbb{P}(dx) \geq 0
\end{equation}

Let \( Q_\epsilon := \left\{ x \in \mathbb{R}^d : \pi_{\text{nr}}(x) \in (\epsilon, 1-\epsilon) \right\} \). The claim (4.10) is implied by the following statement: \( Z(x; \pi_{\text{nr}}) \) is almost surely constant on \( Q_\epsilon \) for each \( \epsilon > 0 \). Assume by contradiction that such constancy does not hold. We then define:

\[
Q_+ := \{ x \in Q_\epsilon : Z(x; \pi_{\text{nr}}) \geq z_0 \}, \quad Q_- := \{ x \in Q_\epsilon : Z(x; \pi_{\text{nr}}) < z_0 \}
\]

Define \( \pi(x) \) as:

\[
\pi(x) =
\begin{cases}
\pi_{\text{nr}}(x) - p_\epsilon & \text{if } x \in Q_+, \\
\pi_{\text{nr}}(x) + p_\epsilon & \text{if } x \in Q_-, \\
\pi_{\text{nr}}(x) & \text{otherwise.}
\end{cases}
\]

It is easy to check that \( \pi \) is feasible and:

\[
J(\pi; \pi_{\text{nr}}) = -p_\epsilon \int_{Q_+} Z(x; \pi_{\text{nr}}) \mathbb{P}(dx) + p_\epsilon \int_{Q_-} Z(x; \pi_{\text{nr}}) \mathbb{P}(dx)
\]

which leads to a contradiction with Eq. \ref{contra}. Finally, we conclude that the stated form of \( \rho_{\text{nr}} \) follows from Eq. \ref{result}.
\hfill $\displaystyle \square$

\subsection{Proof of Theorem 5.1}
\label{supp-ssec:theorem_proof}

Since the assumption states that the empirical risk is a logarithmic function, we can apply Bartlett's second identity so that all $\mH$ is now $\mG$ in the calculation of $Z$. However, a deep learning model's $\mG$ is often not invertible as $n \ll D$. By using damping Hessian \citep{damping} that regularizes the Hessian with a very small number as the dimension of the matrix added to be invertible, this problem can be solved. Also, to deal with many layers of a deep learning model, one can define the model as $f_{\theta}(x) = f_{\theta_L} \circ \cdots \circ f_{\theta_1}(x)$ according to \citep{grosse}. Then, one can show that each layer, $\mG_l(\theta) := n^{-1} \sum_{i=1}^n \nabla_{\theta_l} \ell_i \nabla_{\theta_l} \ell_i^T
$ when $\mG$ is now $\text{diag}(\mG_1(\theta), \dots, \mG_L(\theta))
$ for $l \in \{1,2, \dots, L\}$.

Combining both approaches, $\mG$, the second moment of first gradients, becomes $\left(\mG_l(\theta^*) + \lambda_l I_{d_l}\right).$ Recall that 
\begin{equation}
\mG = \mathbb{E}[\nabla_\theta \mathcal{L}(\theta; \vy, \vx) \nabla_\theta \mathcal{L}(\theta; \vy, \vx)^\top \vert x] \approx \left( \frac{1}{n} \sum_{i=1}^{n} \nabla_{\theta_l} \ell_i \nabla_{\theta_l} \ell_i^T + \lambda_l I_{d_l} \right).
\end{equation}
Then, by the damping factor and Taylor's approximation, the summation can be swapped and by Sherman Morrison's formula, the following holds:
\begin{equation}
\mG^{-1} = (\frac{1}{n} \sum_{i=1}^{n} \left( \nabla_{\theta_l} \ell_i \nabla_{\theta_l} \ell_i^T + \lambda_l I_{d_l} \right))^{-1} = \frac{1}{n \lambda_l} \sum_{i=1}^{n} \left( I_{d_l} - \frac{\nabla_{\theta_l} \ell_i \nabla_{\theta_l} \ell_i^T}{\lambda_l + \nabla_{\theta_l} \ell_i^T \nabla_{\theta_l} \ell_i} \right).
\end{equation}
This derivation is from \citep{datainf} and concludes the time complexity for $\mG = O(nDL)$. Now, by the fact that $n, L \ll D$, the time complexity for $Z$ is also bounded by $O(nDL).$ \hfill $\displaystyle \square$

\subsection{Justification of Surrogate Model using Minimax Theorem.}
\label{supp-ssec:minimax}
First, assume the worst case of the surrogate model, 
\begin{equation}
\sup_\vy\inf_{\Theta^{A}}R_N(\vtheta^{\text{bat}|\text{A}}; \vy).
\end{equation}
Then, from the proof of Proposition~\ref{prop:p1}, we know that 
\begin{equation}
\sup_\vy R(\hat\vtheta^{\text{bat}|\text{A}}_{\text{surrogate}}; \vy) \le \sup_{\vy}\inf_{\Theta^A}R(\vtheta^A; \vy) + \epsilon,
\end{equation}
for some surrogate model with a well designed adaptation. By the assumptions from Sec.~\ref{ssec:assumptions}, Minimax theorem can be applied so that a single worst case comparison can be generalized as
\begin{equation}
\sup_\vy R(\hat\vtheta^{\text{bat}|\text{A}}_{\text{surrogate}}; \vy) \le \inf_{\Theta^A}\sup_{\vy}R(\vtheta^A; \vy) + \epsilon = \sup_{\vy}R(\vtheta^{A^*}; \vy) + \epsilon.
\end{equation}
\hfill $\displaystyle \square$

\section{Experimental Details}
\label{setting}

In this section, we provide detailed explanations of the experimental setups and methodologies used in our study. Our experiments involve both diffusion model and language model to validate the propositions and evaluate the performance of various algorithms.  \\
For the diffusion model (DreamBooth and LyCORIS), we used the LAION dataset \citep{laion} as the backbone dataset $\mathcal{D}^\text{B}$, since Stable Diffusion \citep{ldm} is pre-trained on it. We gathered adaptation datasets $\mathcal{D}^\text{A}$ from sources like Textual Inversion \citep{text_inversion} and Kaggle's ``Star Wars" dataset \citep{starwars}. For the language model, we employed LLaMA 2-7B-alpaca-cleaned as the backbone language model. This model is LLaMA 2-7B~\citep{llama2} specifically fine-tuned on the Alpaca-cleaned dataset~\citep{alpaca-cleaned}. Since most language models do not disclose their pre-training datasets, we adopted this publicly available model that had undergone further fine-tuning.

\textbf{DreamBooth.} 
For DreamBooth, all training was performed using a single NVIDIA RTX4090 GPU per adaptation. The typical learning rate was 5e-6. We used the AdamW optimizer  for the entire training, with $\beta_1 = 0.9$ and $\beta_2 = 0.999$, a weight decay of 1e-2 andepsilon set to 1e-8. All inference seeds began with 42 and increased by 1 for each loop. \\
We gathered adaptation datasets from Textual Inversion \citep{text_inversion}, consisting of 5 images (e.g., red teapot and elephant datasets). DreamBooth's own dog dataset was also composed of 5 images. To construct the experiments, we generated strong surrogate and BAT models with 800 steps. BAT datasets were created by adding LAION data to the original datasets, and BAT training was conducted with these datasets.

\textbf{LyCORIS.}
The LoCon algorithm, part of the LyCORIS library, introduces a low-rank adaptation technique specifically designed for convolutional layers in diffusion models like Stable Diffusion. Our experiments were conducted based on Stable Diffusion 1.4 as the backbone diffusion model~\citep{ldm}. Originally developed by~\citep{lora} for attention layers in large language models, this adaptation for convolutional layers enhances image quality and fidelity during fine-tuning. For parameter-efficient fine-tuning (PEFT), we utilized LoCon among the LyCORIS methods. The learning rate was set to $5 \times 10^{-6}$, and the optimizer used was AdamW with $\beta_1 = 0.9$ and $\beta_2 = 0.999$. All training steps were fixed at 200, and a subset of these steps was plotted.

The dataset consists of movie character images sourced from a public dataset available on Kaggle, specifically the `Star Wars' dataset~\citep{starwars}. Among the datasets used during the experiments applying LyCORIS PEFT, we focused on the characters Admiral Piett, Bodhi Rook, and Rose Tico. To train the optimal model and the BAT algorithm, we used different numbers of images per character. The optimal models for Admiral Piett and Bodhi Rook were trained on 91 images each, and Rose Tico's optimal model utilized 94 images. In contrast, the BAT algorithm used fewer images—10 for Admiral Piett, 43 for Bodhi Rook, and 38 for Rose Tico. When obtaining benchmark scores, we retrained the models with 300 training steps, keeping other experimental settings the same, and saved the model every 50 steps to extract the scores.

\textbf{LoRA \& DoRA.} 
For LLaMA 2 based adaptations, NVIDIA A6000 GPUs are used according to the required experiments. LoRA's rank was set to 8. LoRA alpha was 32, and dropout was given by 0.1. Target model was query and value matrices of each transformer layer. The learning rate was 5e-5, and normally the batch size was 64. Weight decay was set to 0.01. 
We took MedQuad \citep{medquad}, WinoGrande \citep{winogrande}, and XSum \citep{xsum} as adaptation datasets $\mathcal{D}^\text{A}$. To build the BAT set $\mathcal{D}^{\text{bat}|\text{A}}$, we sampled $\mathcal{D}^\text{B}$ at regular intervals and inserted the samples into $\mathcal{D}^\text{A}$, also at regular intervals. Here, we set $|\mathcal{D}^\text{A}|=10000$ as a default.

\textbf{Backbone Augmentation Ratio}. The backbone augmentation ratio is fixed at $\gamma = 0.95$ for DreamBooth and LoCon, meaning that 5\% of the original adaptation data is replaced with backbone data. For LoRA and DoRA, we set $\gamma = 0.9$. Exceptionally, for the HellaSwag benchmark, we set $\gamma = 0.992$ for DoRA and $\gamma = 0.9993$ for LoRA, respectively.

\section{Additional Experiments}
\label{ablation}

\subsection{Metrics} \label{benchmark} Using DINOv2~\citep{dino}, cosine similarity is used to measure the similarity between two feature vectors, often extracted from image representations. Given two vectors \( \mathbf{v}_1 \) and \( \mathbf{v}_2 \), their cosine similarity is computed as:
\begin{equation}
\text{Cosine Similarity}(\mathbf{v}_1, \mathbf{v}_2) = \frac{\mathbf{v}_1 \cdot \mathbf{v}_2}{\|\mathbf{v}_1\| \|\mathbf{v}_2\|}.
\end{equation}
The centroid represents the mean vector of a set of feature vectors. The squared centroid is the square of the distance between the centroid and each data point. Suppose we have \( N \) data points \( \mathbf{v}_i \in \mathbb{R}^d \). The centroid \( \mathbf{c} \) is given by:
\begin{equation}
\mathbf{c} = \frac{1}{N} \sum_{i=1}^N \mathbf{v}_i.
\end{equation}
The squared centroid distance for each point \( \mathbf{v}_i \) is:
\begin{equation}
\text{Squared Centroid Distance} = \sum_{i=1}^N \|\mathbf{v}_i - \mathbf{c}\|^2.
\end{equation}
Where \( \|\mathbf{v}_i - \mathbf{c}\|^2 \) is the squared Euclidean distance between each point and the centroid. Lower centroid score shows that the output is more consistent with lower variance which infers better generalization.

CLIP uses cosine similarity to compare text and image embeddings. The model learns to maximize the similarity between matching text-image pairs while minimizing the similarity between non-matching pairs.
Let \( \mathbf{t} \) be the text embedding and \( \mathbf{i} \) be the image embedding. The similarity score between them is calculated as:
\begin{equation}
\text{CLIP Similarity}(\mathbf{t}, \mathbf{i}) = \frac{\mathbf{t} \cdot \mathbf{i}}{\|\mathbf{t}\| \|\mathbf{i}\|}.
\end{equation}
As \( \mathbf{t} \cdot \mathbf{i} \) is the dot product between the text and image embedding, and \( \|\mathbf{t}\| \) and \( \|\mathbf{i}\| \) are the norms of the text and image embeddings. The cosine similarity is maximized for relevant text-image pairs and minimized for irrelevant pairs.

The Vendi score is a metric used to quantify similarity across multiple domains or datasets. It measures the overlap between sets of embeddings from different modalities (e.g., vision, text). Mathematically, Vendi score uses the concept of overlapping support across distributions.

Given two distributions of feature vectors \( P \) and \( Q \), the Vendi score can be formulated as:
\begin{equation}
\text{Vendi Score}(P, Q) = \int \min(P(x), Q(x)) dx.
\end{equation}
This score evaluates how much of the support of one distribution is shared by the other, effectively measuring their similarity. Higher Vendi scores indicate greater overlap between distributions. Therefore, in the case of adaptations, lower Vendi scores implies the concentration of identity.

\clearpage

\subsection{More Quantitative BAT Results}
\label{more_quan}

\begin{table}[h!]
\centering

\caption{\textbf{Comparison of MMD and FID for Medical MRI Synthesis.} To challenge our method even further, we adopted data from different domain. MRI data from Mendeley Lumbar Spine Dataset is used in this experiment. 10\% of the data is utilized and $\gamma = 0.95.$. The results report that even in extreme domain shift, BAT retains its effectiveness.}

\begin{tabular}{|l|c|c|}
\hline
Metric & \textbf{MMD} $\downarrow$ & \textbf{FID} $\downarrow$ \\
\hline
Baseline & 0.489 & 36.43\\
\textbf{Ours} & \textbf{0.408} & \textbf{34.71}\\
\hline
\end{tabular}
\end{table}

\vspace{-3pt}

\begin{table}[h!]
\centering

\caption{\textbf{Comparison between Selected Data and Original Data.} Comparison of Cosine SIM and CLIP Text scores. $\gamma$ = 0.6 for DreamBooth Data is used and $\gamma = 0.95$ for Star Wars data. The results show that the BAT chosen data is more similar in both text and image embedding spaces to the original data than randomly selected data.}

\begin{tabular}{|l|c|c|}
\hline
Score & Cosine Sim $\uparrow$ & CLIP Text $\uparrow$ \\
\hline
DreamBooth (Random) & 0.4655 & 0.6788 \\
\textbf{Ours} & \textbf{0.6557} & \textbf{0.6895} \\
Star Wars (Random) & 0.4021 & 0.4518 \\
\textbf{Ours} & \textbf{0.4605} & \textbf{0.6865} \\
\hline
\end{tabular}
\end{table}

\vspace{-3pt}

\begin{table}[h!]
\centering

\caption{\textbf{FID $\downarrow$ Comparison of performance metrics for DreamBooth and Star Wars Datasets.} Since FID is one of the most important metrics in image generation, we measured the score of our method and compared it with former methods in various settings. The result shows that our method generates better quality images with more genuine characteristics.}

\begin{tabular}{|l|c|c|c|c|}
\hline
\textbf{Models} & \textbf{Baseline} & \textbf{Random} & \textbf{Weak(Ours)} & \textbf{Strong(Ours)} \\
\hline
DreamBooth Dataset & 31.01 & 29.26 & 22.71 & \textbf{20.92} \\
Star Wars Dataset & 80.19 & 79.91 & \textbf{67.43} & 70.22 \\
\hline
\end{tabular}

\end{table}

\vspace{-3pt}

\begin{table}[h!]
\centering

\caption{\textbf{Metric Results of Distance-based Data Selection and Ours.} We compared data selection method used in fine-tuning method and BAT. The results conclude that our method is more effective in data selection.}

\begin{tabular}{|l|c|c|c|c|}
\hline
Score & Cosine Sim $\uparrow$ & Centroid Distance $\downarrow$ & CLIP $\uparrow$ & Vendi $\downarrow$ \\
\hline
\rowcolors{2}{gray!50}{white} 
Ours (DreamBooth Data) & \textbf{0.8959} & \textbf{102.8} & 0.2501 & \textbf{1.258} \\
Distance-based & 0.8492 & 148.6962 & \textbf{0.2531} & 1.577 \\
\rowcolors{2}{gray}{white}
Ours (Star Wars Data) & \textbf{0.4494} & \textbf{590.1} & \textbf{0.2598} & \textbf{7.583} \\
Distance-based & 0.4394 & 602.9975 & 0.2497 & 7.924 \\
\hline
\end{tabular}
\end{table}

\vspace{-3pt}

\begin{table}[h!]
\centering

\caption{\textbf{Metric Results for Less Backbone, Full Backbone, and BAT.} Regarding the comparison of the addition of backbone or new adaptation data, we conducted an experiment with personalization adaptation using Star Wars dataset by comparing the results from 95\% of adaptation data, 100\% adaptation data and 95 \% adaptation and 5 \% BAT data. The result is shown below.}

\begin{tabular}{|l|c|c|c|c|}
\hline
Score & Cosine Sim $\uparrow$ & Centroid Distance $\downarrow$ & CLIP $\uparrow$ & Vendi $\downarrow$ \\
\hline
Less Backbone & 0.4106 & 628.3 & 0.2571 & 8.461 \\
Full Backbone & 0.4326 & 613.1 & \textbf{0.2715} & 8.047 \\
BAT & \textbf{0.4494} & \textbf{590.1} & 0.2598 & \textbf{7.583} \\
\hline
\end{tabular}

\end{table}

\vspace{-3pt}

\begin{table}[h!]
\centering

\caption{\textbf{Metric Results for BAT with COCO Dataset.} We adopted another backbone dataset to show the consistency of BAT. The following table justifies the claim.}

\begin{tabular}{|l|c|c|c|c|}
\hline
Score & Cosine Sim $\uparrow$ & Centroid Distance $\downarrow$ & CLIP $\uparrow$ & Vendi $\downarrow$ \\
\hline
DreamBooth Dataset & 0.8501 & 149.3 & \textbf{0.2519} & 1.358 \\
Ours (COCO) & \textbf{0.8959} & \textbf{102.8} & 0.2501 & \textbf{1.258} \\
Star Wars Dataset & 0.4106 & 628.3 & 0.2571 & 8.461 \\
Ours (COCO) & \textbf{0.4494} & \textbf{590.1} & \textbf{0.2598} & \textbf{7.583} \\
\hline
\end{tabular}

\end{table}

\vspace{-3pt}

\begin{table}[t!]
\centering

\caption{\textbf{Metric Results for Applying ALBAT for Synthetic Datasets.} We applied ALBAT to the synthetic data of personalization adaptation which is a common practice. The table shows BAT can be also effective in synthetic data as well.}

\begin{tabular}{|l|c|c|c|c|}
\hline
Score & Cosine Sim $\uparrow$ & Centroid Distance $\downarrow$ & CLIP $\uparrow$ & Vendi $\downarrow$ \\
\hline
DreamBooth Dataset & 0.8501 & 149.3 & \textbf{0.2519} & \textbf{1.358} \\
Synthetic BAT & \textbf{0.8843} & \textbf{113.5} & 0.2421 & 1.371 \\
Star Wars Dataset & 0.4106 & 628.3 & 0.2571 & 8.461 \\
Synthetic BAT & \textbf{0.4525} & \textbf{593.1} & \textbf{0.2741} & \textbf{7.186} \\
\hline
\end{tabular}
\end{table}

\clearpage

\subsection{More Qualitative BAT Results}
\label{sub-experiments}
\begin{figure}[h!]
    \label{fig:qualitative_score1}
    \vskip 0.2in
    \begin{center}
    \centerline{\includegraphics[width=\linewidth]{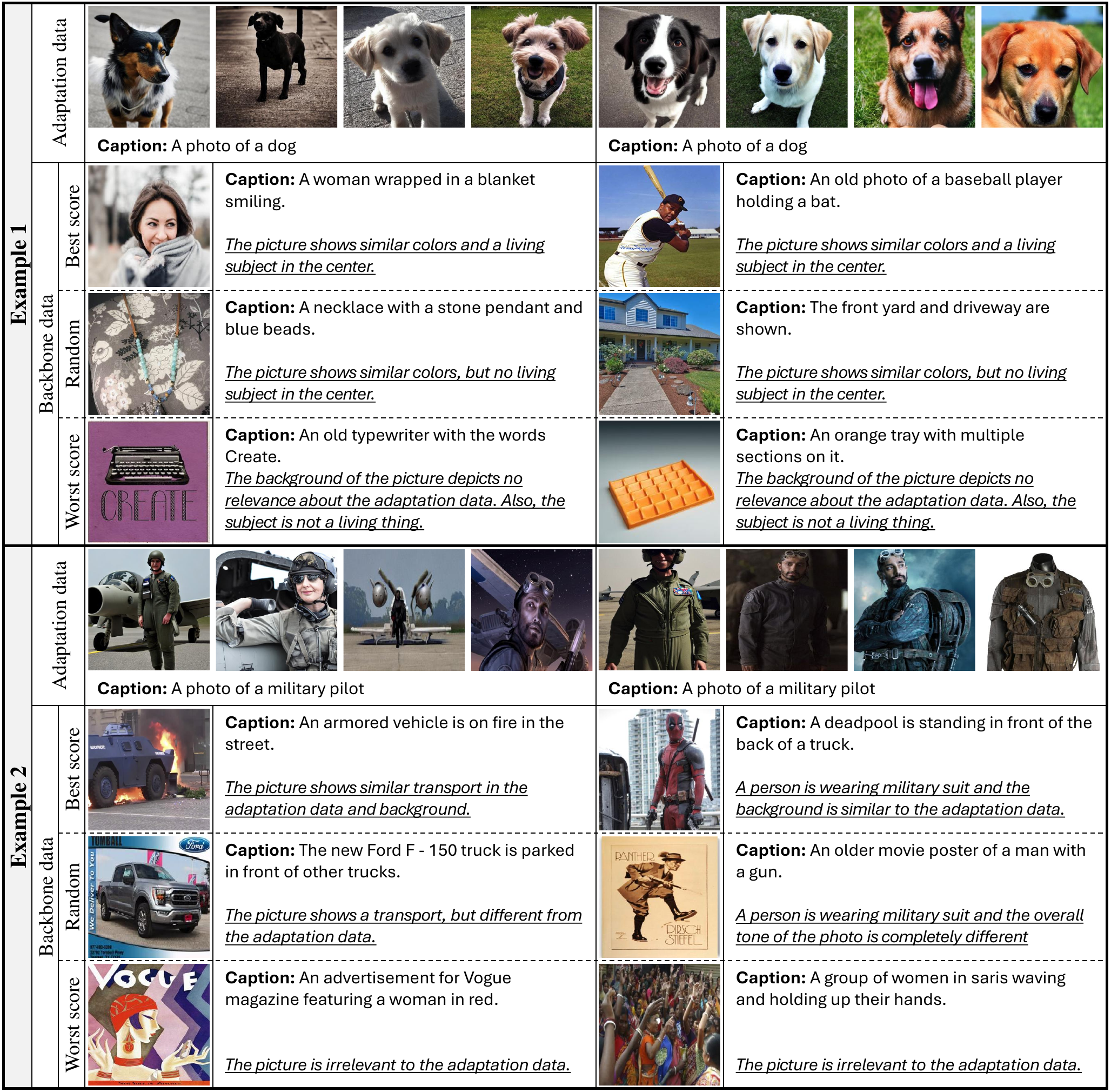}}
    \caption{
    \textbf{\batalg{} Results for personalization task.} Given adapation data, these are the backbone data (LAION dataset) examples retrieved from \batalg{}. Since we have added regularization data from DreamBooth as adaptation data as well, \batalg{} has chosen data that are similar to the regularization data mostly. The selected data points show some intuitive understanding of the benefit that these added backbone data will give to the augmented adaptation training. The first and second row of each example are the top two and the worst two BAT data selection results.}
    \end{center}
    \vskip -0.2in
\end{figure}

\begin{figure}[h]
    \label{fig:qualitative_score2}
    \vskip 0.2in
    \begin{center}
    \centerline{\includegraphics[width=0.86\linewidth]{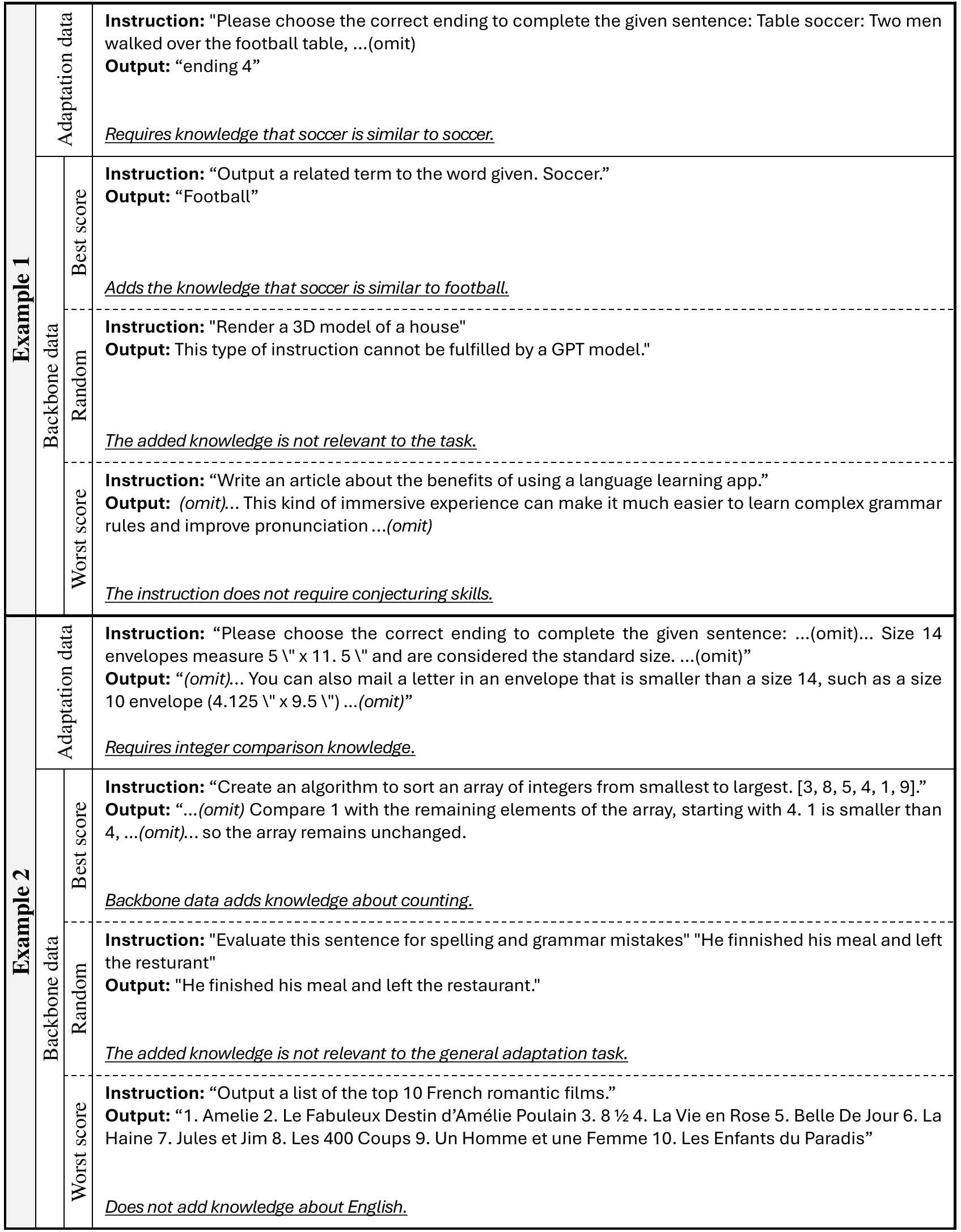}}
    \caption{
    \textbf{\batalg{} Results of commonsense reasoning task.} These are the backbone data examples retrieved from \batalg{}. Best score data add conducive knowledge to adaptation tasks but worst score data do not show such behavior.
    }
    \end{center}
    \vskip -0.2in
\end{figure}

\begin{figure}[h!]
    \label{fig:qualitative_coco}
    \vskip 0.2in
    \begin{center}
    \centerline{\includegraphics[width=\linewidth]{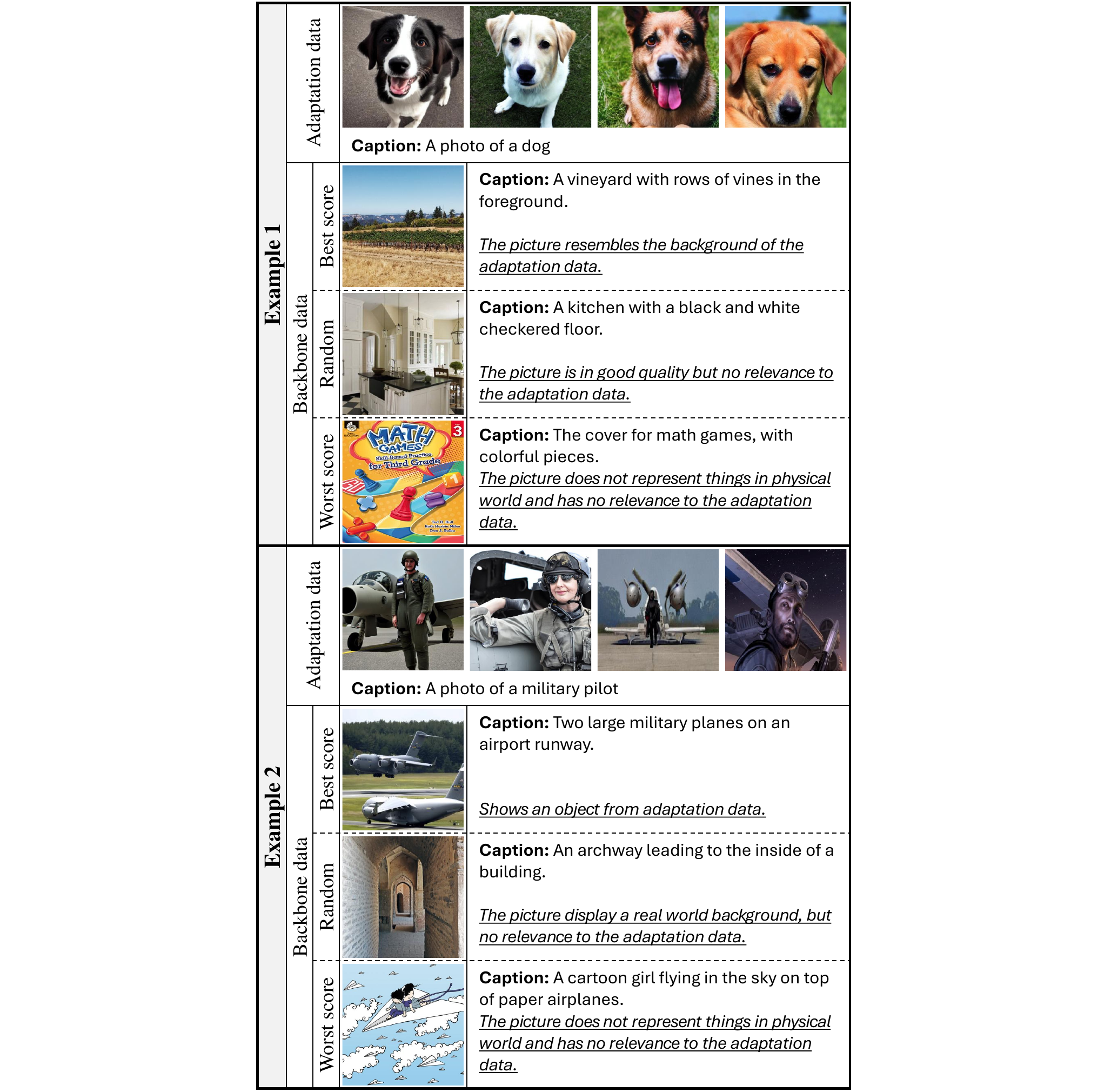}}
    \caption{
    \textbf{ALBAT with COCO Dataset.} This is the same experiment as in Fig.~4 in Supp.~Sec~\ref{sub-experiments}, but the COCO dataset was used as the backbone dataset.
    }
    \end{center}
    \vskip -0.2in
\end{figure}

\begin{figure}
    \vskip 0.2in
    \begin{center}
    \centerline{\includegraphics[width=\linewidth]{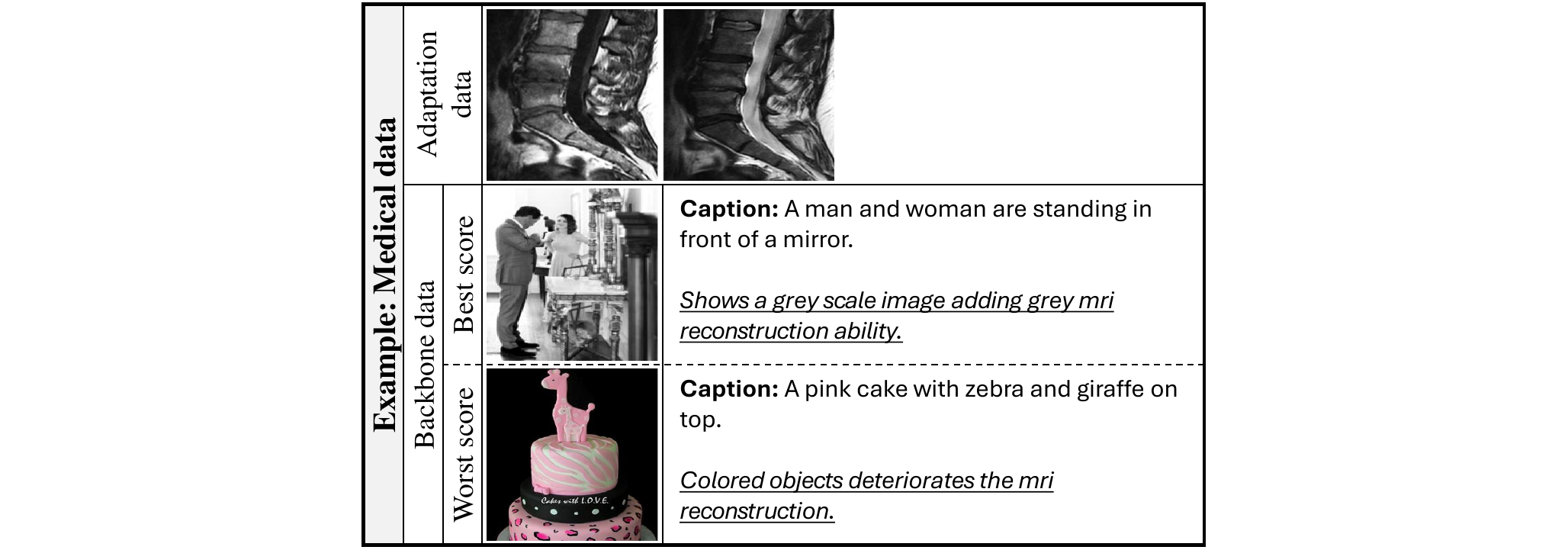}}
    \caption{
    \textbf{ALBAT Selection on Spine MRI Synthesis.} The setting of this experiment is the same as that in Fig.~5 of Supp.~Sec~\ref{sub-experiments}, except that the adaptation dataset used is Mendeley Spine MRI Synthesis. The results of this experiment show that even when the domains differ greatly, ALBAT is still capable of selecting backbone data that is beneficial for training.
    }
    \end{center}
    \vskip -0.2in
\end{figure}

\begin{figure}
    \vskip 0.2in
    \begin{center}
    \centerline{\includegraphics[width=\linewidth]{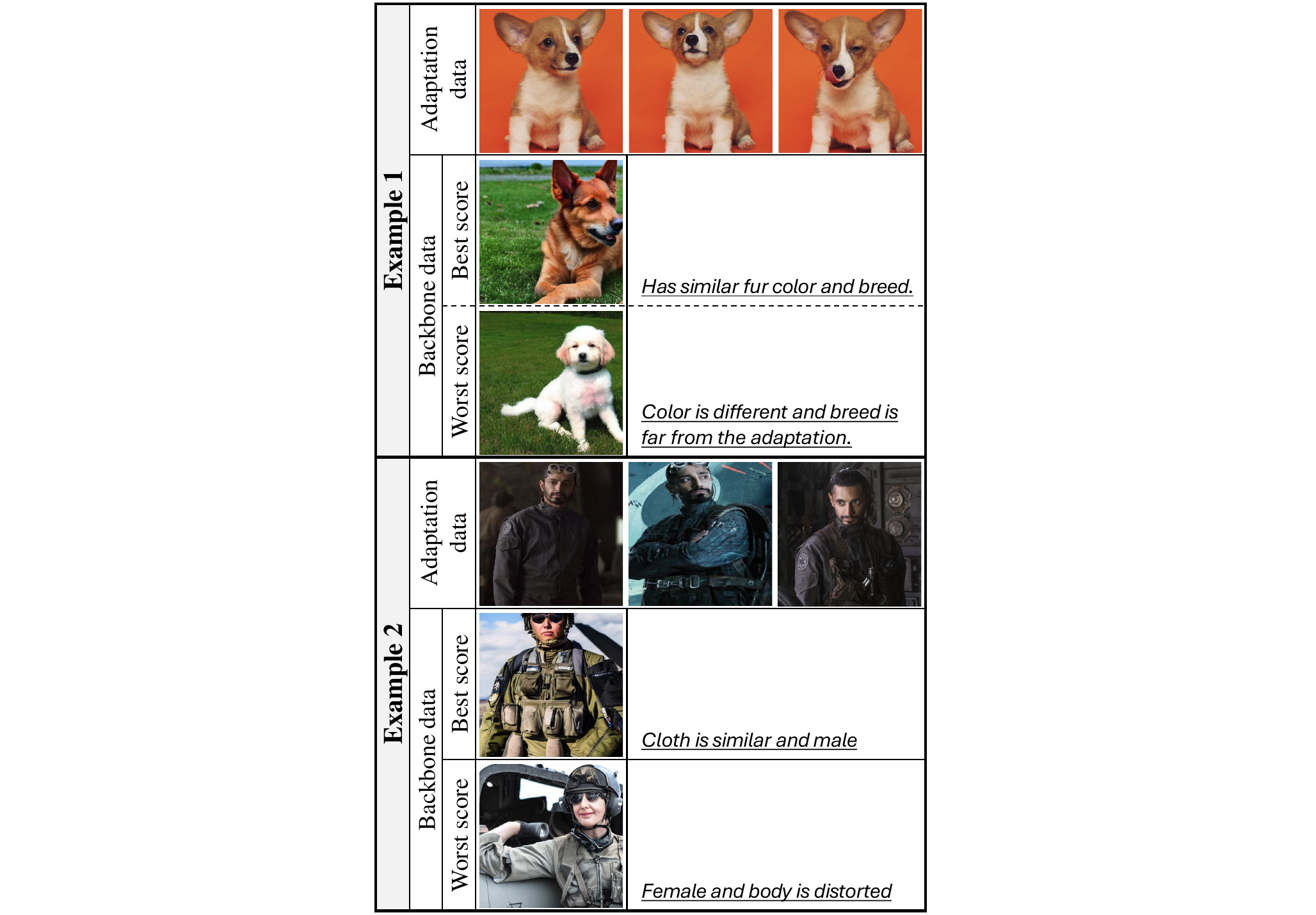}}
    \caption{
    \textbf{ALBAT Selection on Synthetic Data Augmentation.} This is an ablation result with augmentation data generated by a diffusion model. Pre-trained Stable Diffusion is used to generate 100 data points with the prompt \textit{a photo of dog} and \textit{a photo of military pilot}.
    }
    \end{center}
    \vskip -0.2in
\end{figure}

\begin{figure}[h]
    \label{fig:qualitative}
    \vskip 0.2in
    \begin{center}
    \centerline{\includegraphics[width=0.86\linewidth]{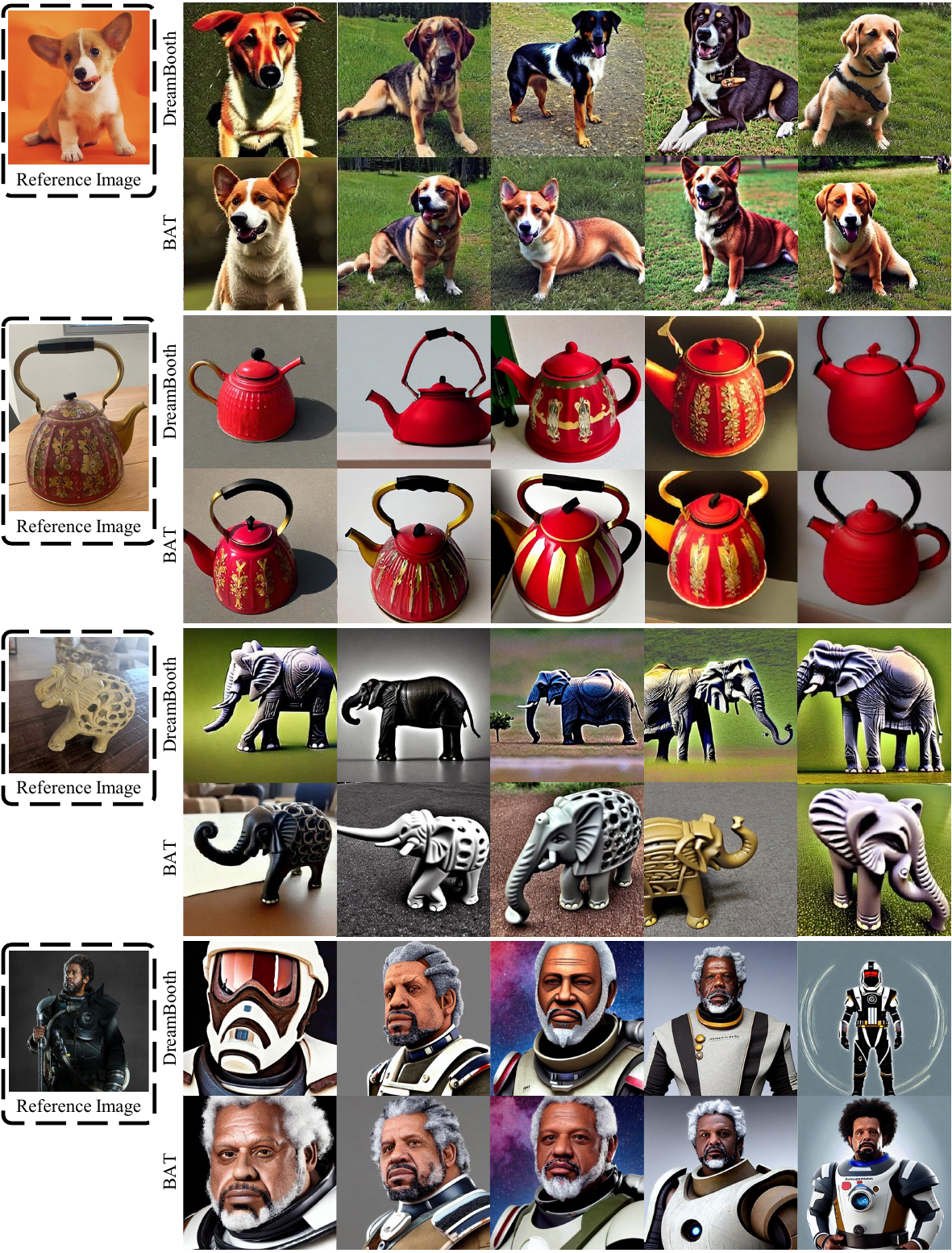}}
    \caption{
    \textbf{DreamBooth Qualitative Outcomes.} These outcomes are gathered from original DreamBooth and BAT. Some of the images do not show acceptable personalization outcome as our experiments assume limited adaptation data setting. This experiment has limited the training steps to 400, half of the original adaptation training. Every class uses the same backbone model and each set of the compared images is generated using an identical setting. Compared to the original adaptation, BAT shows better personalization results.
    }
    \end{center}
    \vskip -0.2in
\end{figure}

\end{document}